# Scaling In, Not Up? Testing Thick Citation Context Analysis with GPT-5 and Fragile Prompts

Arno Simons, Technische Universität Berlin

## Abstract

This paper tests whether large language models (LLMs) can support interpretative citation context analysis (CCA) by scaling in thick, text-grounded readings of a single hard case rather than scaling up typological labels. It foregrounds prompt-sensitivity analysis as a methodological issue by varying prompt scaffolding and framing in a balanced 2×3 design. Using footnote 6 in Chubin and Moitra (1975) and Gilbert's (1977) reconstruction as a probe, I implement a two-stage GPT-5 pipeline: a citation-text-only surface classification and expectation pass, followed by cross-document interpretative reconstruction using the citing and cited full texts. Across 90 reconstructions, the model produces 450 distinct hypotheses. Close reading and inductive coding identify 21 recurring interpretative moves, and linear probability models estimate how prompt choices shift their frequencies and lexical repertoire. GPT-5's surface pass is highly stable, consistently classifying the citation as "supplementary". In reconstruction, the model generates a structured space of plausible alternatives, but scaffolding and examples redistribute attention and vocabulary, sometimes toward strained readings. Relative to Gilbert, GPT-5 detects the same textual hinges yet more often resolves them as lineage and positioning than as admonishment. The study outlines opportunities and risks of using LLMs as guided co-analysts for inspectable, contestable interpretative CCA, and it shows that prompt scaffolding and framing systematically tilt which plausible readings and vocabularies the model foregrounds.

## Keywords

Large Language Models (LLMs), Prompt-Sensitivity Analysis, Citation Context Analysis, Sociology of Science, Interpretative Methods

## 1. Introduction

First introduced nearly fifty years ago in *Social Studies of Science* by Moravcsik and Murugesan (1975) and Chubin and Moitra (1975), citation context analysis (CCA) has been used to specify what in-text references are doing in their immediate context, often so that citation counts and maps of science can be interpreted more cautiously (cf. Bornmann and Daniel, 2008; Kunnath et al., 2021). This ambition has most often been pursued in a scaling-up mode: build comprehensive classification schemes, then develop automated classifiers that can apply them across large corpora. Recent transformer-based models, including fine-tuned encoder models and generative large language models (LLMs), have renewed this scaling-up ambition by making automated classification easier to build and easier to use (e.g., Beltagy et al., 2019; Kunnath et al., 2021; 2023; Nishikawa and Koshiba, 2024; Zhang et al., 2023)

A fundamental critique of scaling up CCA was articulated just as early, most prominently by Gilbert (1977). For classification to travel at scale, readers must be expected to converge on what a reference is doing. Gilbert challenges that expectation by arguing that referencing is often a form of persuasion, directed toward particular audiences in particular situations, so that citation meaning may not be read uniformly. Reviews of citing behavior likewise point to plural motivations, ranging from credit-giving to strategic positioning, and note a further complication for CCA: negative assessments are often voiced indirectly or disguised rather than marked by explicit "negational" cues (e.g., Iqbal et al., 2021; Kunnath et al., 2021; Tahamtam and Bornmann, 2019). Typological CCA's scalability therefore rests on a sociologically fragile premise: that reasonably competent readers can "recognize references as instances of one or other of their categories" and also "share a common reading of the significance and meaning of references" (ibid, p. 119).

This concern also connects to a broader hermeneutic obstacle. As Cronin (1998, p. 48) notes, "a full understanding of why *A* cites *B* requires a multi-layered explanation and, ideally, thick description of the process — and the politics of the process". Such description presupposes familiarity with the case and its technical and social terrain. As Cozzens (1985, p. 135) puts it, one "needs to know a good deal about the scientific area being studied" before the materials make sense, and even Moravcsik and Murugesan (1975, p. 87) pointed to the limits faced by analysts who are "not equipped to understand the technical scientific content" of the papers they code, a jab aimed at "sociologists" and "people in the library sciences". Even with expertise, however, the upshot is not only that some readings are better than others, but that interpretative plurality is to some extent irreducible, partly because authors themselves leave meanings ambiguous or voice negative assessments indirectly. As Gilbert (1977) and others (e.g., Law and Williams, 1982) emphasize, analyses of scientists' writing practices rely on standpoints that can rarely be finally defended and can at best claim pragmatic force.

These long-standing objections become newly pressing now that LLMs are widely taken to perform something like contextual interpretation, at least in the limited sense of generating plausible readings on demand. Until recently, a common assumption was that automation could, at best, approximate surface-level classifications of citation contexts (e.g., Teufel et al., 2006; White, 2004), and that interpretation as an "ill-structured" activity (Gläser and Laudel, 2013, p. 2) lay beyond algorithmic capture. White (2004, p. 103) captured this skepticism sharply when he argued that schemes requiring close reading, domain knowledge, expert judgment, and the recovery of implicit meaning "cannot be delegated to a computer even in principle".

Against this background, the aim of this paper is not to use LLMs to scale CCA *up*, but to test whether they can help scale interpretative work *in*. Rather than treating disagreement and multiplicity as noise to be eliminated in the name of a single stable label, I treat them as part of the object and ask whether a prompted model can generate a structured range of text-grounded readings for a single hard case. I use OpenAI's GPT-5 model as a guided co-analyst: it is made to state a surface-level classification first, then to check expectations against the citing and cited texts, attend to lexical and rhetorical cues, and finally propose multiple distinct interpretative hypotheses that can be compared and contested. Taking the idea of interpretative plurality seriously, I do not assume that there is a single "correct" reading that a better prompt would simply recover. If citation meaning varies with standpoint, audience, and the analyst's reconstruction of context, then different ways of instructing the model should also shift what it treats as salient and which readings it generates. I therefore vary prompt structure and framing systematically and examine how those variations redistribute the interpretations the model produces.

The paper is structured as follows. Section 2 introduces footnote 6 in Chubin and Moitra (1975) as a hard case for citation context analysis and reconstructs Gilbert's (1977) interpretative path through it. Section 3 translates that path into a two-stage prompting pipeline and reports the 2×3 prompt-variation design, run plan, and coding plus modeling strategy. Section 4 presents results from the surface-level classification baseline and from the interpretative reconstructions, then tests how scaffolding and prompt framing redistribute the model's hypotheses and vocabulary, and finally compares GPT-5's cue handling and hypotheses to Gilbert's trajectory. Section 5 discusses what these findings imply for "scaling in" interpretative work with LLMs, including methodological opportunities, risks, and limitations, and outlines directions for applying the workflow beyond this single case.

## 2. Footnote 6 as a hard case

To test what it would mean to scale interpretative CCA in rather than scaling classification up, I return to a canonical hard case from the field's early debates: footnote 6 in Chubin and Moitra (1975) and Gilbert's (1977) reading of it. The note is brief, but it concentrates several of the interpretative problems discussed above, including audience effects, the underdetermination of "function" labels, and the possibility that negative assessment is voiced indirectly rather than signaled by explicit cues. The footnote reads:

> [6] A distinction between 'references' and 'citations' was introduced by D. J. deS. Price, 'Citation Measures of Hard Science, Soft Science, Technology and Non-Science', in C. Nelson and D. Pollock (eds.), Communication Among Scientists and Engineers (Lexington, Mass.: D. C. Heath, 1970), 3-22, esp. 7, then reiterated by G. N. Gilbert and S. Woolgar, 'The Quantitative Study of Science: An Examination of the Literature', Science Studies, 4 (1974), 279-94. We relax that distinction until the findings of our analyses are reported. (Chubin and Moitra, 1975, p. 424)

On a surface reading, this looks like routine attribution and housekeeping. Price is credited with introducing the distinction, Gilbert and Woolgar with repeating it, and Chubin and Moitra announce that they will temporarily set it aside for the purposes of their analysis. Nothing appears overtly polemical.

Gilbert's (1977) treatment of the footnote shows why that appearance is not decisive. He reconstructs the case as a sequence of interpretative moves. First, he treats the footnote as an instance for Chubin and Moitra's own scheme and notes that it could be coded as either "perfunctory" or "additional", but that neither choice "seems self-evidently appropriate" (1977, p. 120). Second, he focuses on the wording, especially "reiterated", and asks why Gilbert and Woolgar are cited at all if they merely repeat Price. Third, he draws on contextual knowledge of the cited works and observes that Gilbert and Woolgar (1974) introduce the distinction without citing Price at the definitional moment. Fourth, he compares that observation with the phrasing "introduced by Price, then reiterated by Gilbert and Woolgar", reading it as a potentially consequential allocation of credit. Finally, he proposes that the first sentence "could… be read as an implied admonishment" for failing to give credit, and that it might therefore merit a negational classification (ibid; emphasis added).

For Gilbert, the point is not to establish the correct label or to claim privileged access to authorial intention. It is to show how quickly classification becomes underdetermined once one takes seriously that intentions are not usually available to the analyst and that competent readers may disagree, especially when they lack insider familiarity with the relevant texts and disputes. In this sense, footnote 6 functions less as an oddity than as a worked demonstration of how interpretative reconstruction can destabilize what looked, at first glance, like a neutral citation.

This is also why footnote 6 has been treated as exemplary in the CCA literature. Swales (1986) singled it out as one of the rare cases where typologies were illustrated with actual citation text rather than category names alone, and he used it to underline how elusive negational references can be. Such references are often consequential for questions of influence and priority, yet they may be easy to miss precisely because disagreement is voiced indirectly or embedded in what otherwise reads as attribution. Later discussions of annotation difficulty and disagreement in citation-function coding have returned to similar issues, including the limits of local cues, praise that conceals critique, scope decisions about what counts as "context", and the practical challenges of resolving references across documents (Iqbal et al., 2021; Kunnath et al., 2021; Tahamtan and Bornmann, 2019).

At the same time, Gilbert's handling of footnote 6 can be read as a template for interpretative CCA. His reconstruction begins with a surface-level coding attempt, then questions its adequacy, isolates potentially loaded wording, checks expectations against the cited and citing texts, and only then proposes an alternative reading. The value of this sequence is not that it delivers a definitive verdict, but that it generates a more articulated space of plausible readings and makes visible the evidential hinges on which those readings turn.

That is the role footnote 6 plays in this paper. It serves as a probe for whether an LLM can be guided through a comparable sequence and, in doing so, contribute candidate reconstructions that remain anchored in textual features while widening the range of interpretations available for scrutiny. The goal is not to replace interpretative judgment, but to test whether some of the labor of producing and organizing alternative readings can be delegated to a prompted model in a controlled way, and to examine what is gained and lost when it is.

The analysis is organized around five research questions that follow directly from this hard case and from Gilbert's critique. First, I ask how stable GPT-5's surface-level classification of footnote 6 is when it is constrained to operate in a typological, scaling-up mode (RQ1). Second, once the model is given the relevant full texts and procedural scaffolding, I ask what kinds of alternative interpretations it generates and how varied they are (RQ2). Third, taking seriously the possibility that interpretative outputs depend on how the task is posed, I test whether changes in prompt structure and framing shift the distribution of interpretations the model produces (RQ3). Fourth, I use Gilbert's admonishment trajectory as a historically salient point of comparison, asking how close the model comes to that reading and where it systematically diverges (RQ4). Finally, I draw these results together to assess what the experiments suggest about using LLMs to scale interpretative work in rather than scaling classification up, including both opportunities and methodological risks (RQ5).

## 3. Design and methods

This section describes the two-stage prompting pipeline, the 2×3 prompt-variation design, the run plan, and the analytic strategy used to assess variation in GPT-5's outputs.

### 3.1 Reconstructing and operationalizing Gilbert's path

Starting from the premise that Gilbert's analysis of footnote 6 can serve as a procedural model for computational interpretative CCA, this subsection translates his interpretative moves into discrete subtasks that an LLM can perform. I implemented this as a two-stage pipeline in which one instance of GPT-5 produces an initial surface-level analysis and a second instance carries out the interpretative reconstruction.

In stage one, the model completed two tasks on footnote 6. First, it classified the citation using the six mutually exclusive, hierarchical categories proposed by Chubin and Moitra (1975): *Essential-Basic*, *Essential-Subsidiary*, *Supplementary-Additional-Information*, *Supplementary-Perfunctory*, *Negational-Partial*, and *Negational-Total* (see Supplementary Section S1 for an explanation of the scheme). Second, it formulated expectations about the likely content of the cited sources, including their substantive themes and whether they would themselves reference particular works. In both tasks the model was instructed to rely strictly on the explicit wording of the citation context and to avoid speculation about hidden motives or implicit critique. This stage was designed to approximate the kind of surface-level classification that scientometric typologies typically aim for (Teufel et al., 2006). The model's input was a slightly simplified version of footnote 6 with normalized formatting: "A distinction between 'references' and 'citations' was introduced by Price (1970), then reiterated by Gilbert and Woolgar (1974). We relax that distinction until the findings of our analyses are reported".

In stage two, a new model instance received the full stage-one outputs (classification and expectations) together with the full-text PDFs of the citing work, Chubin and Moitra (1975; hereafter C&M), and the cited sources, Price (1970) and Gilbert and Woolgar (1974; hereafter G&W). It was instructed to move beyond the surface label and explore alternative meanings or functions the citation might serve. To do so, it followed a four-step procedure that mirrors the remainder of Gilbert's path: (1) expectation checks against the cited texts, noting confirmations or mismatches with brief quotations, (2) cue analysis of lexical or rhetorical features that might suggest meanings beyond a straightforward classification, (3) extended-context analysis of how footnote 6 functions within C&M (placement, recurrence, co-citation relations, and narrative role), and (4) hypothesis generation. In the final step the model produced five distinct hypothesis–justification pairs. Each hypothesis was required to complicate the surface-level classification, be grounded in textual cues from the citing and/or cited texts, engage with evidence from steps 1–3, and use a different angle of reasoning than the others.

The pipeline combines prompt chaining and stepwise prompting. The shift from stage one to stage two is a chained transition, in which the output of the first call becomes input to a new model instance, preventing later interpretative work from influencing the initial classification. Within each stage, tasks were elicited stepwise within a single call. This choice supports the paper's aim of generating multiple, articulated readings under explicit constraints rather than maximizing throughput. It also mitigates a known risk in multi-step prompts: models can sometimes produce strategically weak early outputs to set up a stronger "discovery" later (Sun et al., 2024). By separating the surface-level pass from the interpretative reconstruction across two calls, the design reduces that incentive at the key boundary. Given these tradeoffs, I treat intermediate step outputs as heuristic windows on model behavior rather than as a faithful trace of sequential reasoning.

## 3.2 Experimental configurations

Having outlined the two-stage prompting framework, I now describe how it was implemented under systematically varied conditions. If interpretative CCA depends on how an analyst—or here, an LLM—is guided through the material, then even small changes in prompt design or framing may shift the readings that emerge. To probe this sensitivity, I varied the stage-two prompt across six configurations in a balanced 2×3 design: two base prompt structures crossed with three nudging conditions.

The two base prompts (reproduced in full in Supplementary Section S2) varied the degree of scaffolding:

> **4-step base prompt**: the stage-two procedure described above, with explicit outputs for expectation checks, cue analysis, extended-context analysis, and hypothesis generation.
>
> **1-step base prompt**: the model received the same stage-one outputs and PDFs but was instructed to move directly to five textually anchored hypothesis–justification pairs, without intermediate steps. The hypotheses still had to complicate the surface-level classification, be grounded in textual cues from the citing and/or cited texts, and differ in angle from one another.

The base prompts were varied to test whether comparable interpretative depth is possible without explicit scaffolding. I nonetheless expected the 4-step prompt to steer the model more toward specific hinges in this case—such as "reiterated" and the (non-)citation of Price in G&W—and therefore to approach Gilbert's admonishment trajectory more readily.

The three nudging conditions were implemented by adding, or not adding, a short paragraph of examples in the "Role and Objective" section at the start of the stage-two prompt:

> ***No-nudge***: No extra paragraph was added.
>
> **Nudging *Toward* the specific case of footnote 6**: The following paragraph was added:
> For example, a neutral reference can mask a corrective undertone, hinting at overlooked connections or antecedents. By highlighting some works and downplaying others, a paper can rewrite an idea's genealogy. A nod to a critical source may mute that critique, acknowledging it only to push it aside. Citations can also legitimize shortcuts, suggesting that methodological slippage is acceptable because others have done it. And by citing across divides, authors may enroll rival audiences while reframing their perspectives as compatible with their own.
>
> **Nudging *Away from* the specific case of footnote 6**: The following paragraph was added:
> For example, overly generous praise can mask underlying criticism, serving as a veiled form of ridicule, where excess flattery highlights the weakness of the work. An otherwise unnecessary reference may act as a nod to collaborators, reinforcing social ties rather than substantive claims. A cluster of citations can lend borrowed authority even when the sources are only loosely connected. Long reference lists may project encyclopedic coverage without real engagement. And some references work less for peers than for students, pointing readers toward introductory or survey material.

The aim of these nudges was to test susceptibility to example-driven steering. Examples in prompts can improve specificity, but they can also seed storyline templates and vocabulary that narrow the space of readings. The *Toward* nudge was intended to make available a family of interpretations that, for this case, strike me as relatively plausible—lineage or priority work, muted critique, audience bridging, and the strategic legitimation of temporarily relaxing the ref–cit distinction—without channeling a single Gilbert-like conclusion. The *Away* nudge, by contrast, foregrounded interpretations I judged less plausible here—praise-as-ridicule, purely collegial nods, borrowed-authority clusters, and didactic onboarding—in order to stress-test how readily the model follows an offered framing even when it fits the case poorly. Taking interpretative plurality as part of the object rather than noise, the broader goal was to test how such illustrative examples redistribute that plurality: which plausible readings the model produces more often, which it produces less often, and which vocabulary it uses to articulate them. Because the purpose of the study is to widen, rather than narrow, the interpretative space, the *No-nudge* condition and the requirement to produce multiple distinct hypotheses function as counterweights to any single framing.

## 3.3 Execution and data

All experiments were scripted in Python, which sent prompts to OpenAI's GPT-5 via the API and collected outputs for analysis. The model was run in a high-deliberation ("high-reasoning") configuration, and prompts were issued in batches with responses captured in structured formats for counting and comparison.

Stage one was executed 30 times to assess stability in the surface-level C&M classification of footnote 6 and in the associated expectation notes. Because stage one consistently produced a "supplementary" classification (see Section 4.1), I randomly sampled 15 stage-one outputs to seed stage two.

Stage two implemented the six prompt settings of the balanced 2×3 design (two base prompts × three nudging orientations). Each setting was run once for each of the 15 sampled stage-one outputs, yielding 90 runs in total. Stage two generated 450 hypothesis–justification pairs (90 runs × 5 hypotheses) and, in the 4-step settings, 135 intermediate outputs (45 runs × 3 intermediate steps).

Model specs, tokens, and costs are provided in Supplementary Section S3.

## 3.4 Close reading and statistical analysis

All outputs from stages one and two were examined through close reading to assess the range and plausibility of the model's interpretative reasoning along Gilbert's path. The main material comprised 450 stage-two hypothesis–justification pairs, each treated as a single interpretative unit. An initial attempt to assign one mutually exclusive theme per unit proved unworkable, since many units combined several interpretative moves and the hypothesis field alone was often too compressed. I therefore shifted to recurring subthemes that can co-occur within a unit and, through iterative reading, comparison, and refinement (cf. Glaser and Strauss, 1967), developed 21 binary codes to capture them (Table 1). For ease of reading, I refer to hypothesis–justification pairs as "hypotheses" from this point onward, and I mark explicitly when I discuss hypothesis and justification fields separately.

| Code | Description: Hypothesis interprets footnote 6 as… | N | % |
|---|---|---|---|
| *Agile* | Justifying temporary relaxation of the ref–cit distinction and its flexible use | 208 | 46 |
| *Preempt* | Preempting potential criticism about (mis)use of the distinction | 106 | 24 |
| *Aware* | Signaling familiarity with the distinction and its policing | 73 | 16 |
| *MuteGW* | Downplaying G&W's discussion of the distinction | 73 | 16 |
| *Bridge* | Bridging distinct intellectual camps to signal inclusivity or recruit multiple audiences | 68 | 15 |
| *Test* | Recasting the distinction as an empirical question to be examined with data | 45 | 10 |
| *Pragma* | Projecting a pragmatic and empirical stance | 39 | 9 |
| *Agency* | Asserting authority to reshape the distinction on one's own terms | 37 | 8 |
| *Payoff* | Deferring the distinction to foreground empirical results as the paper's payoff | 34 | 8 |
| *UseGW* | Borrowing G&W's authority instrumentally | 33 | 7 |
| *Canon* | Normalizing the distinction as consensus through a tidy genealogy | 27 | 6 |
| *SSS* | Positioning the paper for the SSS journal, appealing to its reviewers' expectations | 27 | 6 |
| *SideP* | Aligning with Price's bibliometric program rather than with sociology of science | 26 | 6 |
| *PrioP* | Emphasizing Price's priority or originality | 25 | 6 |
| *UseP* | Borrowing Price's authority instrumentally | 14 | 3 |
| *NegP* | Challenging Price's discussion of the distinction | 13 | 3 |
| *NegGW* | Challenging G&W's discussion of the distinction | 10 | 2 |
| *NegGEN* | Challenging the discussion of the distinction in general | 10 | 2 |
| *Teach* | Orienting readers to sources primarily for pedagogy | 8 | 2 |
| *MuteGEN* | Downplaying the discussion of the distinction in general | 6 | 1 |
| *MuteP* | Downplaying Price's discussion of the distinction | 5 | 1 |

**Table 1:** 21 Codes derived from close reading and inductive analysis of 450 stage-two hypotheses generated by GPT-5. The codes summarize the roles attributed to the citation context (footnote 6) and are ordered by decreasing frequency.

To connect these qualitative patterns to experimental conditions, I estimated linear probability models (LPMs) for each of the 21 codes. LPM coefficients are directly interpretable as percentage point differences, which facilitates comparison across codes and conditions (Breen et al., 2018). All models follow the balanced 2×3 design, with indicators for base prompt (*1-step* vs *4-step*), nudging (*Toward*, *Away*, *No-nudge*), and their interaction. I cluster standard errors by run because each run yields five hypotheses under the same settings, so observations within a run may not be independent. This leaves point estimates unchanged but yields more conservative inference by accounting for within-run dependence.

For interpretation I focus on average marginal effects (AMEs), expressed as percentage point differences: AME(*4-step*) compares *4-step* to *1-step* averaged over nudges, while AME(*Toward*) and AME(*Away*) compare each nudge to *No-nudge* averaged over base prompts. These estimands match the design-level questions of interest and provide a compact summary of how scaffolding and framing shift the distribution of coded interpretative moves. Where useful, I report selected pairwise contrasts alongside cluster-robust confidence intervals and p-values. Since the LPMs support joint inference, I also report a 5-df omnibus Wald test across the six settings. This global check is background and does not replace the AMEs, which remain the primary estimands. Because it asks a broad "any effect anywhere?" question, it can be underpowered, so a null result does not imply the AMEs are zero.

See Supplementary Section S4 for more details on the motivation and mathematical specification of the LPMs.

I triangulated the LPM results with close readings of stage-one outputs (classifications and expectations) and, in the *4-step* conditions, the intermediate stage-two outputs (expectation checks, cue analyses, and extended-context analyses). These intermediate texts help clarify how particular hypotheses are assembled and where prompt variations appear to redirect attention. Because the *4-step* outputs were elicited within a single stepwise call, they are best treated as heuristic windows on model behavior rather than as a causal trace of sequential reasoning.

# 4. Results

This section proceeds in four steps. First, I report the stage-one surface-level classifications and expectations that GPT-5 assigned to footnote 6 as a baseline "scaling-up" pass, and use these to answer RQ1. Second, I map the range of stage-two hypotheses and introduce the 21 recurring subthemes, treating them as a way to describe the interpretative space the model opens up for this hard case (RQ2). Third, I test how prompt structure and illustrative examples redistribute that space by shifting the frequencies and phrasing of these subthemes, including via "lexical echo" (RQ3). Fourth, I use Gilbert's admonishment trajectory as a historically salient probe to compare how the model picks up and resolves the same textual hinges (RQ4).

## 4.1 Consistent "supplementary" surface-level classification

Stage one yields a stable "supplementary" classification of footnote 6. For the citation of G&W, the model split between *Supplementary-Perfunctory* in 19 of 30 runs (63 percent) and *Supplementary-Additional-Information* in 11 runs (37 percent). In 29 of 30 runs it treated Price and G&W as a single package, assigning both the same label. Only once did it differentiate, tagging Price as *Supplementary-Perfunctory* while giving G&W the slightly weightier *Supplementary-Additional-Information* label, suggesting that "reiterated" can sometimes be read as adding a small increment of informational weight without pushing the citation outside the supplementary range.

The expectation notes are even more uniform. In 29 of 30 runs the model predicted that G&W would themselves cite Price. When it specified the expected function of that Price citation, it was usually *Supplementary-Additional-Information* (23 of 29 cases, 79 percent). A minority framed Price as *Essential-Basic* (5 of 29), and only one as *Supplementary-Perfunctory*, with no *Essential-Subsidiary* or negational expectations. These expectations were stated with little hedging. In a simple regular-expression count over the expectation notes, the model used "expected to cite/reference" in 27 cases and "likely to" in 2, with no instances of "may".[1]

Taken together, stage one reads the footnote as routine attribution and treats "reiterated" as a straightforward cue of acknowledged precedence. The package labeling and confident prediction that G&W will cite Price establish a clean baseline for stage two, where those expectations can be checked against the full texts and the interpretative space can be opened up. This pattern also matches Gilbert's (1977, p. 120) own remark that the note initially looks "additional" or "perfunctory", with possible tensions surfacing only once wider context is brought in. In this sense, stage one answers RQ1 by showing that, under a typological, scaling-up constraint, GPT-5's surface reading is highly stable and remains clustered within C&M's *Supplementary* categories.

## 4.2 A variety of themes in the hypotheses, grounded by textual evidence

Close reading of the 450 hypotheses (that is, the hypothesis–justification pairs) shows both within-run divergence and cross-run recurrence. Within each run, the five hypotheses are consistently different in angle, as required by the prompt. Across runs, those angles nonetheless recur in recognizable families. GPT-5 variously reads footnote 6 as critique management, as a justification for temporarily loosening conceptual discipline, as signaling awareness or authority, as staging an empirical test, as constructing a genealogy of the ref–cit distinction, as borrowing or muting particular authorities, or as bridging audiences. Because many hypotheses bundle more than one move at once, the analysis that follows focuses on the 21 co-occurring subthemes captured by the binary codes in Table 1.

| # | Hypothesis (interpretation of footnote 6) | Broader theme | Subthemes (Codes) | Setting |
|---|---|---|---|---|
| 1 | **The footnote serves as <u>a prophylactic against terminological critique</u>, <u>authorizing the authors to merge reference-level content coding with citation counts</u> without being accused of conceptual slippage:** The phrasing "We relax that distinction" coupled with immediate reliance on both content analysis of references (pp. 426–434) and SCI citation counts (pp. 430–434) <u>suggests deliberate boundary-blurring</u> to support a hybrid design. | Preempting critique to justify conceptual slippage | *Preempt, Agile, Agency* | *4-step No-nudge* |

[1] The exact regular expressions were "[Mm]ay (reference|cite)", "[Ll]ikely to (reference|cite)", and "[Ee]xpected to (reference|cite)".

| | | | | |
|---|---|---|---|---|
| 2 | **Implicit critique of strict definitionalism in favor of empiricism:** The verb 'relax' and the decision to defer the distinction suggest a gentle pushback against what G&W call for ('we must distinguish carefully,' p. 284). The authors imply that strictness can impede exploratory work and that empirical results will warrant re-imposition later (p. 436). | Empiricisms over definitionalism | *NegGW, Pragma, Test* | *4-step Toward* |
| 3 | **Staging an empirical challenge to the necessity of the distinction to buttress their conclusion about counts.:** “We relax that distinction until the findings of our analyses are reported” implies results will adjudicate whether the distinction matters. In the Summary they say the paper “utiliz[es] the distinction between references and citations” and still argues “we cannot afford to abandon citation counts.” The opening footnote sets up a narrative in which their data are meant to recalibrate the perceived strictness or practical import of the Price/G&W rule. | Staging an empirical test to reinforce their conclusion | *Test, Payoff* | *1-step Away* |
| 4 | **Casting Gilbert & Woolgar as ‘reiterating’ Price subtly recasts their critical review as endorsing a Pricean distinction, strengthening the appearance of consensus.** G&W (1974) do define the terms (p. 284) but cite Price (1970) in that section mainly for Price’s Index (p. 285). C&M’s wording (“reiterated by…”) smooths over G&W’s broader critique of quantitative practices, thereby leveraging their authority to normalize the distinction within C&M’s framework. | Creating the appearance of consensus | *Canon, MuteGW, UseGW, Agile* | *4-step No-nudge* |
| 5 | **Social bridging across constituencies to secure legitimacy with multiple audiences:** Pairing Price (bibliometrics/information science) with Gilbert & Woolgar (sociology-of-science critique) in one breath signals fluency in both camps relevant to Social Studies of Science’s readership. The announced “relaxation” allows the authors to speak flexibly to both—retaining Price’s metrics later (SCI citation counts) while acknowledging sociological sensitivities about categories—thus functioning as audience-alignment rather than mere attribution. | Bridging SSS communities | *Bridge, Aware, SSS, Agile* | *1-step No-nudge* |
| 6 | **Genealogical rewriting that situates Price as foundational and domesticate critiques.** The phrasing “introduced… then reiterated” elevates Price as originator and subsumes G&W within his lineage. Given G&W’s skepticism about descriptive fitting and their own insistence on the ref/cite distinction (pp. 283–286), this move stabilizes the paper’s intellectual ancestry and underplays potential tensions with their approach | Genealogical rewriting to elevate Price | *PrioP, MuteGW, Canon, Preempt* | *4-step Toward* |
| 7 | **The citation to Price selectively extracts definitional authority while tacitly distancing the paper from Price’s broader program (Price’s Index and hard/soft/technology/nonscience taxonomy), which the authors neither adopt nor test:** The reference targets Price (1970) “esp. 7,” the page where he codifies the reference–citation terminology, but Chubin & Moitra do not compute Price’s Index or engage his hard/soft distinctions in their analysis of high-energy physics. Instead, they develop a reference typology and then relate it to citation counts. The footnote thus borrows Price’s definitional capital while sidestepping commitment to his larger, contested framework. | Using Price’s authority without engagement | *UseP, NegP, MuteP* | *1-step Toward* |
| 8 | **Didactic signposting for non-specialist readers:** Pointing to Price (1970, esp. p. 7) and G&W (1974) directs readers to canonical definitions while economizing space. The move functions as a pointer for readers new to bibliometric terminology, not merely attribution. | Didactic signposting | *Teach* | *4-step Away* |

**Table 2:** Eight of the 450 stage-two hypotheses generated by GPT-5, illustrating how broader interpretative themes arise from different combinations of coded subthemes. In each row, the **bold** text is the original stage-two hypothesis, followed by its justification in plain text. Underlining marks the spans used to ground the assigned codes, listed in order of appearance.

Table 2 illustrates how broader themes arise from different combinations of subthemes. Example 1 reads the footnote as a preemptive shield against terminological criticism (*Preempt*) that licenses pragmatic boundary-blurring between reference coding and citation counts (*Agile*), with an added note of authorial authority (*Agency*). Example 2 turns "relax" into a soft challenge to definitional policing in G&W (*NegGW*), framing the move as exploratory empiricism (*Pragma*) and as an empirical question rather than a rule to enforce (*Test*). Example 3 similarly emphasizes Test, but ties it more directly to the paper's argumentative payoff about why citation counts remain valuable (*Payoff*). Example 4 treats "reiterated" as an alignment device that manufactures the appearance of consensus (*Canon*) by downplaying G&W's critical stance (*MuteGW*) while redeploying their standing (*UseGW*). Example 5 reads the paired citations as audience work, signaling fluency across bibliometric and STS constituencies (*Bridge*, *Aware*, *SSS*) while justifying flexible use of the distinction (*Agile*). Example 6 pushes the genealogy angle further, elevating Price's priority (*PrioP*) and subordinating G&W (*MuteGW*), stabilizing ancestry (*Canon*) while anticipating possible objections (*Preempt*). Examples 7 and 8 mark two edge readings in the sample: one that leans toward challenging or sidestepping Price's broader program (*UseP*, *NegP*, *MuteP*), and one that foregrounds a didactic pointer function (*Teach*).

These eight examples serve as illustrative anchor points for mapping a spectrum of interpretative distance that also appears across the full set of 450 hypotheses. Examples 1–6 stay closer to the center of the plausible space, elaborating moves that are well anchored in salient wording ("relax", "reiterated", "esp. 7") and in the paper's structure, that is, how the ref–cit distinction is handled across methods, results, and summary. Other hypotheses remain textually anchored but become harder to sustain once the article is read as a whole. Example 7, for instance, treats footnote 6 as borrowing Price's definitional capital while tacitly distancing the paper from his broader program. I take this to sit near the edge of what is plausible, given how extensively C&M rely on Price elsewhere (four citations to three of his works), how explicitly the note frames the move as a pragmatic loosening rather than a repudiation, and the fact that C&M later claim in their summary to have used Price's distinction. Example 8 similarly pushes toward the edge by foregrounding a didactic signposting function that is not obviously prominent in the surrounding argument. That "center-to-edge" spread is analytically useful in this paper's scaling-in frame. The goal is not to compress plurality into a single label, but to see how wide a plausible space of readings the procedure can generate, and where it begins to drift toward reconstructions that informed readers may treat as strained.

The two edge readings in Table 2 also help anticipate the prompt-sensitivity results that follow. Example 7 comes from a *1-step/Toward* run, and across the full set of 450 hypotheses it is the only case that casts footnote 6 as strongly challenging Price's broader program rather than mainly loosening or managing the ref–cit distinction. More generally, *NegP*-centered hypotheses are more common in *1-step* settings (four further cases) than in *4-step* settings (two), and when *NegP* appears elsewhere it is often secondary, bundled with critique of definitional strictness in Price and or G&W. Example 8, by contrast, comes from a *4-step/Away* run, and didactic signposting readings of that kind occur only under *Away*, where the nudge itself makes "references for students" an available storyline. In the scaling-in frame, this suggests that the procedure can widen the space of readings, but that some of that widening reflects how prompts redistribute interpretative attention, including toward reconstructions that remain textually anchored yet feel strained to an informed reader. I return to *NegP* and *Teach* in the next subsection when I quantify these tilts across the 2×3 design.

Viewed as a set, the examples also show why I treat hypotheses as bundles of co-occurring moves rather than as instances of a single underlying "type". The same recurring subthemes are recombined across different broader readings, so hypotheses overlap in codes like *Agile*, *Preempt*, *Canon*, *MuteGW*, *UseGW*, and *Test* rather than sorting into clean bins. This is the practical reason for the 21-code scheme in Table 1, and it bears directly on RQ2: the model's outputs populate a structured but non-discrete space of alternatives, in which interpretative plurality takes the form of recombinable moves.

A further striking feature is how explicitly many hypotheses anchor themselves in textual detail. They routinely cite the footnote's own wording ("relax", "reiterated", "esp. 7") and connect it to the paper's broader organization (how the distinction is invoked across methods, results, and summary). Several also attempt cross-document checks. For instance, Example 4 distinguishes between G&W citing Price at all and citing him for the definitional moment, and uses that mismatch as evidence within the reconstruction. Even when the resulting reading is debatable, this kind of anchoring shows the model doing more than free-associating labels.

Finally, there is a consistent difference in length between *1-step* and *4-step* settings. Total output per run is higher under *4-step* (7,271 vs. 6,469 tokens), but the final hypothesis–justification pairs are shorter (388 vs. 607 characters), mainly because the justifications are briefer (275 vs. 483, while hypotheses are similar at 110 vs. 120). A plausible interpretation is an allocation effect. Based on manual checks of selected probes, the intermediate outputs in the *4-step* condition seem to absorb lengthy quotations, page citations, and micro-arguments, so the final justifications more often condense material that has already been laid out earlier. In *1-step* runs, by contrast, the final justification has to bundle evidence and warrants in one place, which tends to lengthen it. Across the nudge conditions, length differences appear small.

## 4.3 Prompt structure and framing influence interpretations to some extent

Table 3 gives a first, descriptive map of how the 21 subthemes distribute across the six prompt settings. Two patterns stand out even before modeling. First, much of the repertoire is stable: several frequent codes (notably *Preempt*, and to a lesser extent *Agency*, *Aware*, *Payoff*) show only modest variation across cells, which suggests that the procedure tends to return to a common set of interpretative moves regardless of scaffolding or examples. Second, some subthemes show targeted tilts that line up with the design choices. Under the *4-step* scaffold, *Canon*, *PrioP*, and *UseGW* are more common, especially with *Toward*. Under *1-step*, *Pragma* and *Agile* are more common. By nudge, *Toward* concentrates *MuteGW* and modestly increases *Bridge*, while *Away* concentrates *SSS* and produces the only *Teach*-coded hypotheses. Rare codes remain hard to read from raw counts alone, but *NegGEN* and *NegP* appear most often in *1-step*/*Away*, with *Toward* often at or near zero.

| | *4-step* | | | *1-step* | | |
|---|---|---|---|---|---|---|
| Subtheme (Code) | *Toward* | *Away* | *No-nudge* | *Toward* | *Away* | *No-nudge* |
| *Agency* | 7 (9) | 7 (9) | 5 (7) | 7 (9) | 6 (8) | 5 (7) |
| *Agile* | 29 (39) | 30 (40) | 35 (47) | 40 (53) | 33 (44) | 41 (55) |
| *Aware* | 10 (13) | 13 (17) | 13 (17) | 12 (16) | 15 (20) | 10 (13) |
| *Bridge* | 14 (19) | 11 (15) | 10 (13) | 13 (17) | 12 (16) | 8 (11) |
| *Canon* | 10 (10) | 3 (4) | 7 (9) | 5 (7) | 0 (0) | 2 (3) |
| *MuteGEN* | 0 (0) | 2 (3) | 1 (1) | 0 (0) | 3 (4) | 0 (0) |
| *MuteGW* | 18 (24) | 9 (12) | 8 (11) | 15 (20) | 11 (15) | 12 (16) |
| *MuteP* | 2 (3) | 1 (1) | 0 (0) | 1 (1) | 1 (1) | 0 (0) |
| *NegGEN* | 0 (0) | 1 (1) | 3 (4) | 0 (0) | 4 (5) | 2 (3) |
| *NegGW* | 2 (3) | 1 (1) | 1 (1) | 1 (1) | 4 (5) | 1 (1) |
| *NegP* | 0 (0) | 2 (3) | 2 (3) | 2 (3) | 5 (7) | 2 (3) |
| *Payoff* | 8 (11) | 5 (7) | 6 (8) | 5 (7) | 6 (8) | 4 (5) |
| *Pragma* | 2 (3) | 7 (9) | 3 (4) | 8 (11) | 9 (12) | 10 (13) |
| *Preempt* | 16 (21) | 18 (24) | 15 (20) | 21 (28) | 17 (23) | 19 (25) |
| *PrioP* | 9 (12) | 5 (7) | 2 (3) | 3 (4) | 4 (5) | 2 (3) |
| *SSS* | 3 (4) | 8 (11) | 3 (4) | 1 (1) | 10 (13) | 2 (3) |
| *SideP* | 8 (11) | 4 (5) | 5 (7) | 3 (4) | 1 (1) | 5 (7) |
| *Teach* | 0 (0) | 5 (7) | 0 (0) | 0 (0) | 3 (4) | 0 (0) |
| *Test* | 6 (8) | 6 (8) | 11 (15) | 7 (9) | 6 (8) | 9 (12) |
| *UseGW* | 9 (12) | 7 (9) | 7 (9) | 3 (4) | 4 (5) | 3 (4) |
| *UseP* | 3 (4) | 4 (5) | 3 (4) | 2 (3) | 2 (3) | 0 (0) |

**Table 3:** Frequency and relative frequency (%) of the 21 codes across the 2×3 design. Each cell reports the count out of 75 and the corresponding proportion in parentheses. Codes can co-occur, so percentages do not sum to 100% across codes or columns. Base prompts are *4-step* and *1-step*; nudging orientations are *Toward*, *Away*, and *No-nudge*. N = 450 hypotheses overall; code definitions and totals in Table 1.

To move from the suggestive patterns in Table 3 to a more disciplined test, I use the LPMs. They indicate which tilts are robust in the 2×3 design and how scaffolding and examples redistribute interpretative plurality across the 21 subthemes. Averaged across the three nudging conditions (AME(*4-step*)), moving from the *1-step* prompt to the *4-step* prompt makes *Canon* and *UseGW* more likely, by about +6 percentage points (pp) each. At the same time it makes *Pragma* and *Agile* less likely, by about −7 pp and −9 pp. In substantive terms, the scaffolded procedure makes the model less likely to foreground pragmatic and flexible handling of the ref–cit distinction (*Pragma,*

*Agile*) and more likely to foreground genealogical or authority work (*Canon, UseGW*), including readings that treat "introduced… then reiterated" as doing classificatory housekeeping or selectively redeploying G&W's standing. This tilt is compatible with what the scaffold explicitly requires. By forcing the model to isolate cues, check stage-one expectations against the cited texts, and place the footnote in C&M's wider narrative, it supplies prompts for treating specific phrasings as consequential rather than as merely descriptive.

The *Toward* nudge produces a second, more targeted redistribution. Relative to *No-nudge*, and averaged over both base prompts (AME(*Toward*)), *Toward* increases *PrioP* (+5 pp), *MuteGW* (+9 pp), and *Bridge* (+6 pp), while reducing *NegGEN* (−3 pp). This aligns closely with the role of *Toward* in the design. By supplying illustrative examples that foreground lineage, muted critique, and audience-bridging, *Toward* does not so much create a new reading as make particular ones more available and therefore more frequent.

The *Away* nudge tilts the distribution in a different direction. Compared to *No-nudge* and again averaged over base prompts (AME(*Away*)), *Away* increases *SSS* (+9 pp) and *Teach* (+5 pp) while decreasing *Agile* (−9 pp) and *Test* (−5 pp). The *Teach* shift looks like the clearest case of example-driven steering: didactic readings appear only when the prompt itself supplies "references for students" as an interpretative template. The increase in *SSS* is less direct but still plausible given that the *Away* paragraph foregrounds social and audience functions of citing, which may pull attention toward venue and readership alignment rather than toward treating the footnote as methodological license (*Agile*) or as staging an empirical test of the distinction (*Test*).

Two additional cell-level contrasts (not averaged over base prompts) have 95% confidence intervals that exclude 0: *SideP* is higher in *4-step*/*Toward* than in *1-step*/*Toward* (+7 pp) and *1-step*/*Away* (+9 pp), suggesting that scaffolding plus *Toward* framing jointly encourages alignment-with-Price readings. *NegP* is lower in *4-step*/*Toward* than in *1-step*/*Away* (−7 pp), though because *NegP* is rare overall that difference is best treated as suggestive rather than decisive.

The omnibus joint Wald test (5 df) indicates differences across the six settings for *Canon*, *MuteGW*, *NegP*, *NegGEN*, *Pragma*, *SSS*, *SideP*, *Teach*, and *UseP*. For *UseP*, the estimated differences are small (AMEs below 3 pp) and hinge on a single contrast (+5 pp for *4-step*/*Away* vs *1-step*/*No-nudge*), so I interpret them cautiously. Several codes show little evidence of differences beyond baseline variation (*Agency*, *Aware*, *Payoff*, *Preempt*), and some show none at all (*MuteP*, *NegGW*), which reinforces the general picture. Prompt structure and examples do not rewrite the interpretative repertoire. They selectively tilt where the model spends its interpretative effort within that repertoire.

Figure 1 plots all average marginal effects as dot-and-whisker estimates with 95 percent cluster-robust confidence intervals around a zero line, arranged in three panels for AME(*4-step*), AME(*Toward*), and AME(*Away*). This makes it easy to see at a glance which effects differ clearly from zero.

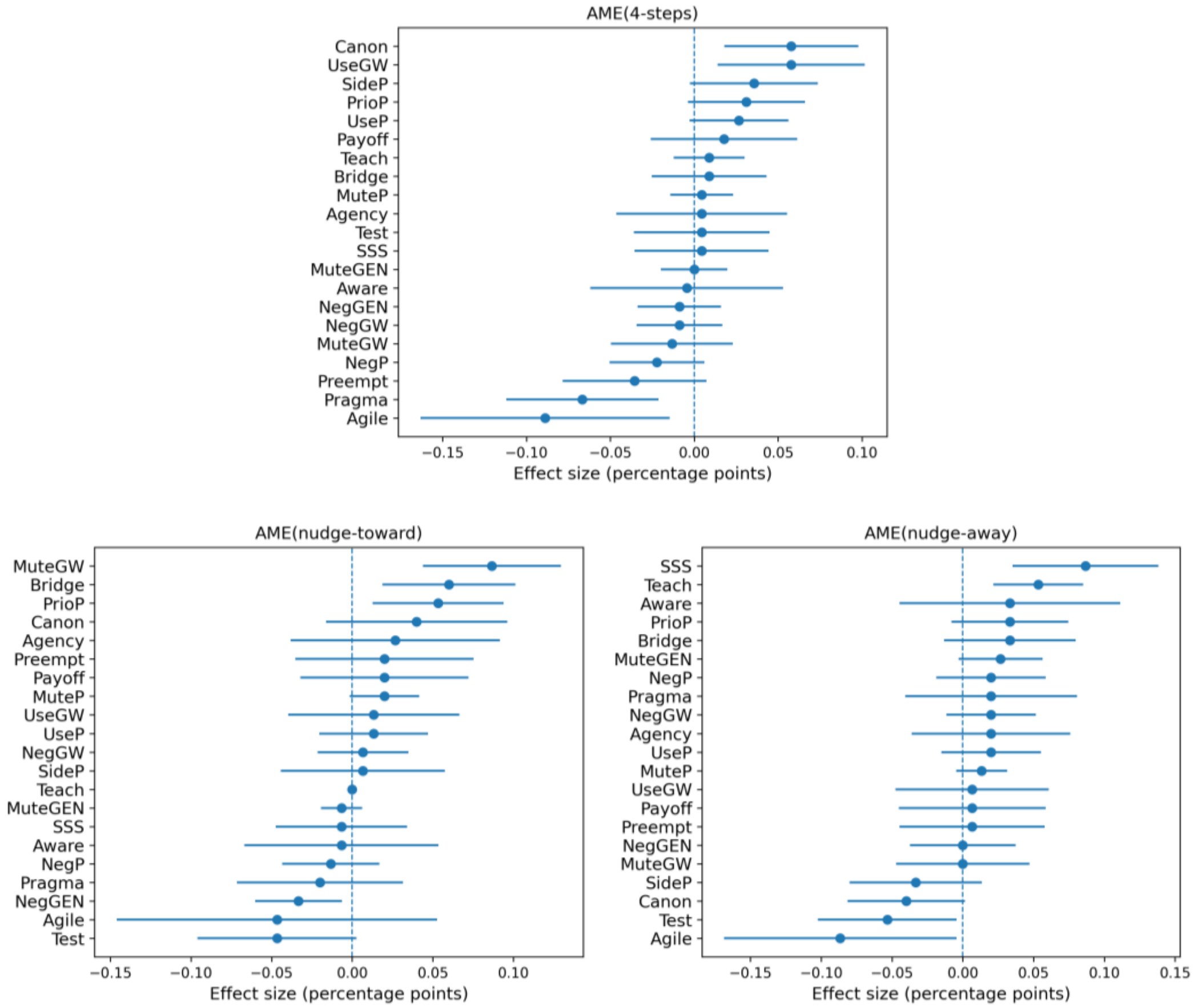

**Figure** 1. Dot-and-whisker plots of average marginal effects (AMEs) for the 21 codes. Each subfigure shows one AME on the x-axis in percentage points for all 21 codes on the y-axis, with dots for estimates and horizontal whiskers for 95% cluster-robust confidence intervals. The vertical line at 0 indicates an effect size of 0. Whiskers that remain entirely on one side of 0 correspond to 95% confidence intervals that exclude 0. Confidence intervals are clustered by run. AME(*4-step*) compares *4-step* to *1-step*, averaged over the three nudges. AME(*Toward*) compares *Toward* to *No-nudge*, averaged over the two base prompts. AME(*Away*) compares *Away* to *No-nudge*, averaged over the two base prompts.

Beyond the subtheme analysis, I also examined “lexical echo”, that is, reuse of vocabulary from the nudging paragraphs in the model’s hypotheses. I lowercased the text, removed stopwords, and then constructed a set of 78 lexical items drawn from the two nudges. For each hypothesis–justification pair, I coded a binary indicator for the presence of each item and fitted the same 2×3 linear probability models as for the codes (for more details and all results see Supplementary Section S5). This analysis shows that *Toward* increases terms such as “genealogy”, “slippage”, “reframing”, and “audience”, while *Away* leaves a weaker and less consistent trace once I adjust for multiple testing. The *4-step* scaffold reduces reuse of generic citation terms shared by both nudges (e.g. “citation”, “reference”), while leaving some more analytic vocabulary intact. Taken together, the prompts shape not only which interpretations are more frequent but also the rhetorical repertoire used to state them.

Taken together, these patterns address RQ3 by showing that prompt design redistributes interpretative plurality within a largely stable repertoire of readings. That matters for the paper's scaling-in framing because the aim is to use the model to generate and organize multiple, text-grounded reconstructions of one hard case. On that view, scaffolding (*4-step* vs. *1-step*) and nudges (*Toward*, *Away*, *No-nudge*) are levers that shift which plausible readings the model foregrounds, which it backgrounds, and what vocabulary it uses to articulate them.

## 4.4 GPT-5 heeds Gilbert's cues but favors muting over admonishment

This subsection treats Gilbert's (1977) reading as a historically salient trajectory through the hard case, not as a ground truth. To my reading, Gilbert's admonishment interpretation is textually well grounded, but it is not the only plausible reconstruction of footnote 6. Price presents the reference–citation distinction as a terminological proposal, and Gilbert and Woolgar echo his language and notation almost verbatim while not crediting him at the definitional sentence, even though they cite the same chapter shortly thereafter for Price's Index and related points. From an attribution perspective, that pattern is underdetermined. It can be read as a meaningful omission or as a minor lapse within an otherwise generous crediting practice. This underdetermination is what gives Chubin and Moitra's phrasing "introduced … then reiterated" its interpretative range: it can read as routine genealogy, it can read as quietly re-centering credit on Price, and it can also support the stronger possibility Gilbert highlights, a veiled rebuke.[2] Against this backdrop, the question is whether GPT-5 tracks the same hinges (see Section 2), and if so, how it tends to resolve them.

Across the 450 stage-two hypotheses, the model never reproduced Gilbert's admonishment reading in full. It did, however, repeatedly converge on the same cue complex that makes that reading possible. The closest hypotheses, including Examples 4 and 6 in Table 2, treat "introduced … then reiterated" as a consequential ordering of authorities that foregrounds Price and potentially diminishes G&W. Where Gilbert turns that mismatch toward a veiled rebuke, the model more often turns it into lineage-management and positioning readings that normalize C&M's move and mute potential conflict.

The *4-step* intermediate outputs (see Section 3.1) give a glimpse of how GPT-5 may have moved from cue detection to hypotheses. In expectation checks, the model almost always noticed that G&W do not cite Price when defining the ref–cit distinction (41 of 45 runs), often adding the sharper qualification that Price is cited there for other

[2] A browser based exchange with ChatGPT-5, reproduced in Supplementary Section S7, follows up on this underdetermination and shows in miniature how an LLM can assemble the relevant passages and spell out arguments on both sides, illustrating its potential as a non-human sparring partner for interpretative CCA.

items (for example Price's Index and related points) rather than as the source of the distinction.

In cue analysis, all 45 runs flagged "introduced … then reiterated" as potentially evaluative. When asked what this might suggest under closer scrutiny, 40 of 45 runs resolved it into hierarchy, genealogy, and strategic downplaying of G&W's critical posture. None resolved it into an implied admonishment. In extended-context analysis it also repeatedly observed that Price is a recurring touchstone in C&M while G&W are not, which is consistent with the text. Where the model was less reliable was in extracting co-citation relations from the PDFs. It repeatedly asserted that C&M co-cited Price with other bibliometric sources such as Merton, Cole and Cole, and MacRae, whereas Price is in fact properly co-cited only with MacRae elsewhere in the paper.

Only three hypotheses explicitly stated Gilbert's most diagnostic hinge, namely that G&W define the distinction without crediting Price at the definitional moment. Besides Example 4 (Table 2), the other two were:

> Recasting the genealogy of the distinction to privilege bibliometrics over sociology: Attribution to Price as introducer and G&W as mere reiterators subtly positions the distinction as originating in bibliometrics. Given that G&W don't credit Price for the distinction in their text (they simply define it), Chubin & Moitra's framing recenters credit on Price and aligns the paper with bibliometric authority while only nominally acknowledging sociological uptake. (Example 9, *4-step/Toward*)

> The authors simplify divergent positions by attributing a stable 'introduced/reiterated' distinction to create the appearance of consensus, thereby making their decision to 'relax' it appear minor and reversible. Price frames his move as a convention he adopts (Price 1970, p. 7–8), and G&W define the terms without attributing them to Price (1974, p. 284). Calling one an 'introduction' and the other a 'reiteration' compresses nuance, smoothing the way for "We relax that distinction…". (Example 10, *4-step/No-nudge*)

In my reading, Example 9 treats "Given that…" plus "recenters credit" as more than a priority claim. It reads C&M as re-centering Price by casting G&W as derivative ("reiteration") and tracing the credit ambiguity to their failure to name him at the definitional point. Even here, though, the thrust is genealogy management and bibliometric alignment, not an explicit chastisement.

Example 10 uses the same mismatch differently. By contrasting "Price frames his move as a convention he adopts" with "G&W define the terms without attributing them to Price", and adding that "[c]alling one an 'introduction' and the other a 'reiteration' compresses nuance", it suggests that C&M's ladder flattens G&W's distinct stance. This nearly implies that the credit relation between Price and G&W needs reconsideration, pulling in the opposite direction to Example 9.

Finally, Example 4 uses the mismatch more lightly. It notes that "G&W (1974) do define the terms (p. 284) but cite Price (1970) in that section mainly for Price's Index (p. 285)", but unlike Examples 9 and 10 the point carries little argumentative weight and reads like an evidential aside. A plausible explanation is procedural: because the prompt required each hypothesis to engage mismatches or cues from earlier steps (see Section 3.1), the model may have mentioned the mismatch mainly to satisfy that anchoring constraint, without integrating it into the core inference.

The above-mentioned overall tendency that the model treats "introduced … reiterated" as potentially evaluative, yet mostly converts that load into lineage and positioning rather than rebuke also appears when we move from the *4-step* prompts to the full set of runs and look at co-occurrence patterns for the stem of the "reiter" cue. When "reiter" is explicitly invoked in a hypothesis, it most often travels with the same "muting and lineage" cluster of moves (especially *MuteGW*, *Canon*, *UseGW*, *PrioP*, and *SideP*), and this clustering looks broadly similar under both base prompts and across all three nudging conditions, with only minor shifts such as *MuteP* joining more clearly under *Away*. I also modeled the mention of "reiter" itself as an outcome using an LPM and prompt effects were present but modest. Overall, this LPM suggests that the *4-step*/*Toward* setting makes the model somewhat more likely to invoke "reiter" explicitly, while the basic co-occurrence cluster around that cue remains broadly similar across prompt structures and nudges. For more details see Supplementary Section S6.

Taken together, these results show that GPT-5 tracks Gilbert's reading closely at the level of cues, yet diverges in how it resolves them. The model repeatedly picks up the same hinges Gilbert used: the "introduced… then reiterated" phrasing and the fact that G&W do not credit Price for the ref–cit distinction. Yet even when hypotheses state this mismatch explicitly, the model never turns it into a veiled admonishment, with Example 9 the closest it comes. Instead, the model reworks the cues into lineage and positioning readings that cast footnote 6 as subsuming G&W under the Price lineage, variously signaling allegiance, glossing over differences, muting G&W's critique, defending C&M's approach, or combining these moves. This answers RQ4. At the same time, close reading of Examples 4, 9, and 10, together with the "reiter" co-occurrence cluster and LPM, offers a glimpse into the model's working: similar cue configurations tend to yield similar muted-lineage readings across prompt structures and nudges, even though *4-step*/*Toward* slightly increases explicit "reiter" use and produces the most Gilbert-adjacent hypotheses. More generally, some hypotheses integrate earlier cues as the argumentative spine, while others mention them only in passing.

## 5. Conclusion and discussion

This paper set out to test an unconventional ambition for computational citation context analysis (CCA). Instead of using an LLM to scale up the classification of what in-text references are doing, it asks what it would mean to scale interpretative work *in*. The scaling-up frame treats citation meaning as something that should converge into stable labels that can travel across corpora. The scaling-in frame treats disagreement, ambiguity, and underdetermination as part of the object, and asks whether an LLM can help generate “thick description” (Cronin, 1998, p. 48) of citation contexts by widening and organizing a space of competing reconstructions across cases. The methodological challenge, then, is how a prompted LLM can be made to expand, structure, and textually anchor that space in a way that remains inspectable, contestable, and usable for sociological interpretation. To address this, I developed a two-stage prompting pipeline and a 2×3 prompt-variation design, which I apply here as a proof of principle to one canonical hard case, footnote 6 in Chubin and Moitra (1975). The case is well suited because it makes the limits of surface labeling unusually visible, and because Gilbert’s (1977) reconstruction provides a concrete interpretative sequence that can be translated into a promptable procedure.

Across the experiments, three implications stand out. First, LLMs can function as a guided co-analyst that helps populate and organize interpretative plurality around a hard case, while leaving final judgement with human readers (cf. Simons et al., forthcoming-a). Second, prompt structure and framing operate as methodologically consequential levers in a scaling-in workflow, because they tilt which regions of the interpretative space are explored more often. Third, translating an interpretative path into a prompt-chained pipeline turns tacit reading routines into explicit components that can be shared, inspected, and experimentally varied, which is especially valuable when the aim is not convergence on one label but a defensible mapping of alternatives.

The findings from RQ1–RQ4 suggest that LLMs can contribute meaningfully to CCA in a scaling-in frame, precisely by not behaving like a scaling-up classifier. In stage one, I separated a scaling-up style surface pass for classification from an initial scaling-in move. GPT-5 produced a remarkably stable surface-level classification of footnote 6 (RQ1), and it also generated expectation notes about how Price and Gilbert and Woolgar were likely to relate to one another based strictly on the citation text, which serve as a baseline for later checks. In stage two, the pipeline deepened the scaling-in procedure by moving from citation-text-only expectations to cross-text testing and interpretative reconstruction. The model was asked to check the stage-one expectations against the full texts and then generate multiple distinct hypotheses that complicate the initial reading. Once it had access to the cited and citing materials, it repeatedly noticed where the stage-one expectations failed, distinguished between Gilbert and Woolgar citing Price in general and not citing him for the ref–cit distinction, and drew on cues such as “introduced…reiterated” and “relax” as well as the paper’s wider argumentative handling of the distinction.

Across runs, this yielded a structured range of hypotheses that were both textually grounded and interpretatively diverse, from readings adjacent to Gilbert’s admonishment trajectory (RQ4) to more adventurous proposals that nonetheless tried to remain evidence-based (RQ2). Many hypotheses fell within what I consider a plausible interpretative band, and some would not have occurred to me on my own. In scaling-in terms, that is the core gain. The pipeline can generate many alternative, textually anchored readings at a stable level of procedural attentiveness, which can counteract the tendency of human analysts to compress interpretative variation once routines become familiar.

On this view, automation's value for interpretive CCA is not faster labeling but assistance with the hardest part of reconstructive work: producing, contrasting, and warranting multiple plausible readings. A model that can generate text-grounded reconstructions on demand can widen the space of candidate interpretations beyond what any single reader would typically produce unaided, while keeping judgement and argumentative responsibility with the analyst. Translating Gilbert's sequence into an explicit procedure also makes hypothesis generation more visible and more reproducible, shifting it from tacit craft to an inspectable workflow. A complementary way LLMs can support interpretative CCA is through interactive "sparring" in a chat interface, where follow-up questions probe warrants, elicit counterreadings, and make plausibility criteria explicit for a case such as footnote 6. The downside is that this dialogic format is harder to standardize and reproduce than a scripted pipeline and is more susceptible to path dependence and user steering (see Supplementary Section S7 for a brief illustration).

A second implication is that the interpretative space opened up by the model is not neutral. The prompt sensitivity analysis shows that design choices about scaffolding and examples shift the distribution of hypotheses (RQ3). The *Toward* and *Away* nudges seeded some storylines and sidelined others. More scaffolding in the 4-step prompt, especially when paired with the case-specific *Toward* nudge, tended to produce hypotheses closer to the center of the plausible interpretative space, often elaborating priority, muted critique, and bridging. The 1-step prompt and the *Away* nudge produced a higher share of edge cases, including readings that depended on less plausible leaps from the textual cues and the paper's argumentative structure.

At the same time, the hypothesis range likely also reflects the model's pre- and posttraining. The fact that GPT-5 never reproduced Gilbert's rebuke interpretation under my scaffolding and examples may not be only a prompt effect. Instruction-tuned GPT-family models are optimised via preference feedback and safety objectives that favour cooperative, non-escalatory continuations, and they have documented agreeableness tendencies (Ouyang et al., 2022; Sharma et al., 2025). It is plausible that the model has a soft bias against conflict-forward reconstructions such as admonishment readings unless prompts explicitly license them, for example by inviting the possibility of veiled reprimand or credit-policing even when the local tone is neutral. More adversarial instructions, or other models with different alignment regimes, might yield a different balance of readings.

I suggest that this sensitivity, both to prompts and to model training, is both a risk and a feature. It is a risk because prompts can amplify the analyst's own blind spots and preferred storylines, and model-level priors can further narrow which regions of the interpretative space are readily explored, which makes explicit documentation of prompt choices and design rationales essential. But it is also a feature because understanding how prompts and model training shape the interpretative space is crucial for improving outputs for a given use case. In the case of footnote 6, scaffolding and case-specific framing improved plausibility in a way that resembles guiding a student or collaborator through a difficult passage. The methodological task is therefore to design prompts that are loose enough to admit surprise and disagreement while specific enough to keep the model within a historically and textually plausible band, and to treat model choice and alignment settings as part of that methodological design rather than as an invisible background condition.

The third implication is reflexive. Translating Gilbert's analysis into a prompt-chained pipeline forced me to spell out interpretative moves that would otherwise remain tacit. These included how to formulate expectation checks, how to search for cues that might destabilize a surface-level label, and how to elicit multiple hypotheses that are genuinely distinct in angle rather than rephrasings of one another. Once such steps are written down, they become available as experimental levers that can be adjusted and recombined. In this sense, the model "prompts back": The need to instruct it pushes toward a more explicit articulation of concepts and reasoning paths, and the sensitivity results provide an empirical picture of where those instructions grip most strongly and where they fade. For interpretative CCA this is an opportunity. LLM based experiments can be used not only to generate candidate readings of citation contexts but also to interrogate and refine the interpretative repertoires that human analysts bring to such readings in the first place.

From a methodological perspective, the study offers a proof of concept for how a scaling-in reading procedure can be translated into a reusable prompting architecture. I instantiated this design by following Gilbert's path through footnote 6, but the two-stage structure is more general. Stage one performs surface-level classification and expectation formulation based on the citation context alone, closely aligned with scaling-up ambitions. Stage two takes those expectations and the full texts as input and guides the model through checks, cue analysis, extended-context analysis, and hypothesis generation, which is aligned with scaling-in ambitions. Because prompts, model settings, and code can be shared, the pipeline turns tacit reading practices into an articulated workflow. By varying base prompts and nudging orientations in a simple 2×3 design and combining close reading with inductive coding and linear probability models, the study also shows how prompt design itself can be treated as an object of analysis rather than an invisible precondition.

The study has clear limitations. First, it centers on a single, well-documented case, so the pipeline is a proof of concept that needs testing on other citation contexts before broader claims are warranted. Second, coding and close reading were carried out by me alone, which means the quantitative contrasts reflect a single reader's interpretative judgements rather than a multi-coder standard. Third, GPT-5 is a proprietary model with opaque training data and internal processes, and I did not compare its outputs to those of other systems, including models with different pre- and posttraining regimes that may populate the interpretative space differently. Fourth, the approach is resource intensive and is best suited to a small number of strategically chosen cases rather than large-scale deployment, which is a deliberate trade-off given the scaling-in aim. Finally, implementing the pipeline requires a degree of LLM literacy, from scripting to prompt design, which may restrict who can apply similar setups in practice (cf. Simons et al., 2025).

With these limitations in view, we can return to White's (2004, p. 103) claim that the recovery of implicit meaning "cannot be delegated to a computer even in principle" and to Gilbert's insistence that his footnote 6 reading requires familiarity with the broader context. The experiments reported here suggest that these claims look less absolute in light of current models, at least when the task is posed in a scaling-in way. GPT-5 could carry out several procedural tasks that expertise often involves, including cross-text expectation checking and generating text-grounded hypotheses about what a citation is doing in context.[3] In that sense, an LLM can stand in for some aspects of expert familiarity by widening the interpretative possibility space and reorganizing interpretative labor. At the same time, the human analyst retains responsibility for deciding which regions of that space to cultivate, which hypotheses to discard, and how to carry promising readings further. This collaborative configuration, with the model as a guided co-analyst and the human reader as curator and critic, is how I answer RQ5. The question is therefore no longer whether contextual interpretation and the recovery of implicit meaning can be automated in principle, but how we choose to combine computational and human capacities in the reconstructive study of citation practices.

## Funding statement

This work was supported by the European Union under an ERC Consolidator Grant (Project No. 101044932, "Network Epistemology in Practice (NEPI)"). Views and opinions expressed are however those of the author only and do not necessarily reflect those of the European Union or the European Research Council. Neither the European Union nor the granting authority can be held responsible for them.

[3] In addition, GPT-5 handled several seemingly more "mundane" subtasks in a strikingly competent way, including basic PDF parsing and passage retrieval. At the same time, some operations proved more fragile, especially tracking co-citations across long texts and footnotes and reconstructing citation networks.

# Supplementary Material

## S1 Chubin and Moitra's classification scheme

Chubin and Moitra (1975) proposed the following six mutually exclusive and hierarchical categories to capture a spectrum of citation functions:

***Essential-Basic***: "The referenced paper is declared central to the reported research, a reference on which its findings depend (e.g. the source of the derivation of a fundamental equation or a detailed description of the experimental conditions)".

***Essential-Subsidiary***: "A specific method, tool, or mathematical result is not directly connected to the subject of the paper, but is still essential to the reported research".

***Supplementary-Additional-Information***: "The referenced paper contains an independent supportive observation (idea or finding) with which the citer agrees".

***Supplementary-Perfunctory***: "Papers referred to as related to the reported research without additional comment".

***Negational-Partial***: "If a citer suggests that the paper is erroneous in part and then offers a correction".

***Negational-Total***: "When the citer refers to the paper as being completely wrong and offers an independent interpretation or solution".

## S2 Base prompts

### 4-step base prompt

# Role and Objective

You are a sociologist of science specializing in interpretative analyses of citation contexts within scientific literature. Your main focus is to understand how scientists use citations not only as acknowledgements but also as rhetorical devices. You operate on the premise that citations are deliberate and strategic acts, often carrying meanings that extend beyond their surface appearance.

You will receive:

- A short passage from a scientific paper (the citing paper) that contains one or more references to other publications (the cited paper(s)).

- A surface-level reading and classification of that citation context.

- Explicit expectations derived from the surface-level reading on both what the cited papers are expected to contain or claim, and which papers they are expected to cite and for what purpose.

- The full text of the citing paper.

- The full text(s) of the cited paper(s).

Your task is to probe beyond the surface-level interpretation and investigate what additional meanings or functions the citation context may serve, using a close comparison between how the cited papers are framed by the citing paper and how they present themselves.

# Instructions

Follow the four interpretive steps below. Your task is to separate factual baseline checks (Step 1) from interpretive analysis (Steps 2–4). While all steps must be anchored in textual evidence, the interpretive steps allow room for speculation as long as it is anchored in explicit textual cues, contrasts, or patterns.

## Step 1: Pedantic Expectation Check

For each cited paper, evaluate the two expectation types (if any are given):

- Content expectation
  - Presence: Is the expected argument, claim, or concept found? ("yes"/"no"/"not applicable")
  - Framing: If present, is it framed as expected? ("expected"/"different"/"not applicable")
- Citation expectation
  - Presence: Does the paper cite the expected source(s)? ("yes"/"no"/"not applicable")
  - Function: If so, is the citation used in the expected rhetorical or epistemic role? ("expected"/"different"/"not applicable")

Instructions

- Support every label with direct quotation(s) or page/section reference(s).
- If present but not as anticipated: set Presence = "yes" and Framing/Function = "different".
- If no expectation was provided, or if the note says something like "nothing to expect", use "not applicable".
- If Presence = "no", set the corresponding Framing/Function to "not applicable".
- Keep nuance in the justification, not in the labels.
- Do not speculate about motives; establish a factual baseline only.

## Step 2: Question the Surface-Level Reading

Building on any mismatches identified in Step 1, critically examine the wording of the citation context itself.

- Identify up to five lexical or rhetorical cues that may suggest meanings beyond a straightforward surface reading.
- For each cue, briefly explain what alternative interpretation it could plausibly signal.
- Exclude cues that do not imply alternative meanings.

## Step 3: Extended Citation Context

Consult the full citing paper to contextualize the citation passage more broadly. All interpretive claims in this step must be grounded in explicit textual cues, contrasts, or patterns.

- Placement: Where in the paper does the citation occur? Is it central or peripheral?

- Recurrence: Are the same sources cited again elsewhere in the paper? If so, how are they treated, and does this shed light on the specific citation context under analysis?

- Relational cues: How do the authors position themselves in relation to the cited papers or authors across the paper?

- Co-citation patterns: Are the cited papers clustered with particular other sources in a way that suggests deeper interpretive significance? If so, what significance might this clustering convey?

- Narrative function: What broader role does the citation play in the paper's overall argument or storyline, beyond its surface-level use?

## Step 4: Offer Five Alternative Interpretive Hypotheses

Drawing on your work in Steps 1–3, propose five distinct hypotheses about what the citation context may be accomplishing beyond its surface-level reading. These hypotheses may be speculative, but only in the form of anchored interpretation: they must be explicitly tied to textual cues or patterns, and any reference to authorial intention should be framed as an inference from such cues rather than as a factual claim.

Each hypothesis should:

- Contradict or complicate the surface-level reading.

- Be grounded in cues from the citing context, the broader citing paper, and/or the cited paper(s).

- Directly engage with mismatches from Step 1, cues from Step 2, or contextual insights from Step 3.

- Use a different angle of reasoning than the other hypotheses. Do not simply restate the same interpretation in different words.

# Input Format

- **Surface-level reading and classification**:

  - A single JSON object (as a string) with:

    - **citation_context** (string): The passage with the citation(s).

    - **citing_paper** (string): Identifier of the citing paper.

    - **cited_papers** (array of objects): Objects for each citation, each containing:

      - **cited_paper** (string): Identifier of the cited paper.

      - **classification_category** (string): The assigned citation category.

      - **classification_explanation** (string): A brief explanation of why this classification was assigned.

      - **content_expectation** (string): Expected argument or content.

      - **citation_expectation** (string): Expected referenced papers and their expected function.

- **Full-texts**:

  - PDF files of the citing paper and the cited paper(s), named exactly as in the **citing_paper** and the **cited_paper** fields, respectively. These are included with the user message.

# Output Format

Return a single JSON object with:

- **expectation_check** (array): One object per cited paper:
  - **cited_paper** (string): Identifier of the cited paper.
  - **content_presence** (string): "yes"/"no"/"not applicable" — is the expected content present?
  - **content_framing** (string): "expected"/"different"/"not applicable" — is it framed as anticipated?
  - **content_justification** (string): Supporting quote(s) or reference(s).
  - **citation_presence** (string): "yes"/"no"/"not applicable" — are the expected citations present?
  - **citation_function** (string): "expected"/"different"/"not applicable" — are they used as anticipated?
  - **citation_justification** (string): Supporting quote(s) or reference(s).
- **lexical_cues** (array): Up to five objects:
  - **cue** (string): Lexical or rhetorical cue.
  - **explanation** (string): Possible alternative meaning it signals.
- **extended_context** (object):
  - **placement** (string): Location in paper; central or peripheral.
  - **recurrence** (string): Whether and how cited papers reappear.
  - **relational_cues** (string): Author stance toward cited works.
  - **co_citation_patterns** (string): Groupings with other sources and their deeper significance.
  - **narrative_function** (string): Role in the paper's overall argument.
- **alternative_hypotheses** (array): Exactly five objects:
  - **hypothesis** (string): Plausible alternative interpretation of the citation context beyond the surface reading.
  - **justification** (string): Brief reasoning or evidence.

# Output Constraints

- Return only a valid JSON object in the exact structure specified.
- Do not include any commentary, explanations, or metadata outside the JSON.

# Validation

- Confirm that the output is a single, well-formed JSON object matching the specified schema.

# Stop Condition

- Finish when all steps are executed and output follows the required schema.

## 1-step base prompt

# Role and Objective

You are a sociologist of science specializing in interpretative analyses of citation contexts within scientific literature. Your main focus is to understand how scientists use citations not only as acknowledgements but also as rhetorical devices. You operate on the premise that citations are deliberate and strategic acts, often carrying meanings that extend beyond their surface appearance.

You will receive:

- A short passage from a scientific paper (the citing paper) that contains one or more references to other publications (the cited paper(s)).

- A surface-level reading and classification of that citation context.

- The full text of the citing paper.

- The full text(s) of the cited paper(s).

Your task is to probe beyond the surface-level interpretation and investigate what additional meanings or functions the citation context may serve, using a close comparison between how the cited papers are framed by the citing paper and how they present themselves.

# Instructions

Carefully analyze both the citing paper and the cited paper(s). Based on this analysis, propose five distinct hypotheses about what the citation context may be accomplishing beyond its surface-level reading and classification. These hypotheses may be speculative, but only in the form of anchored interpretation: they must be explicitly tied to textual cues or patterns, and any reference to authorial intention should be framed as an inference from such cues rather than as a factual claim.

Each hypothesis should:

- Contradict or complicate the surface-level reading.

- Be grounded in cues from the citing context, the broader citing paper, and/or the cited paper(s).

- Use a different angle of reasoning than the other hypotheses. Do not simply restate the same interpretation in different words.

# Input Format

- **Surface-level reading and classification**:

  - A single JSON object (as a string) with:

    - **citation_context** (string): The passage with the citation(s).

    - **citing_paper** (string): Identifier of the citing paper.

    - **cited_papers** (array of objects): Objects for each citation, each containing:

      - **cited_paper** (string): Identifier of the cited paper.

      - **classification_category** (string): The assigned citation category.

      - **classification_explanation** (string): A brief explanation of why this classification was assigned.

- **Full-texts**:

  - PDF files of the citing paper and the cited paper(s), named exactly as in the **citing_paper** and the **cited_paper** fields, respectively. These are included with the user message.

# Output Format

Return a single JSON object with:

- **alternative_hypotheses** (array): Exactly five objects:

  - **hypothesis** (string): Plausible alternative interpretation of the citation context beyond the surface reading.

  - **justification** (string): Brief reasoning or evidence.

# Output Constraints

- Return only a valid JSON object in the exact structure specified.

- Do not include any commentary, explanations, or metadata outside the JSON.

# Validation

- Confirm that the output is a single, well-formed JSON object matching the specified schema.

# Stop Condition

- Finish when all steps are executed and output follows the required schema.

## S3 Model specs, tokens, and costs

All runs used “gpt-5-2025-08-07” with “reasoning_effort=‘high’” and “temperature=1.0”. Other generation parameters were left at their API defaults.

Across all experiments, the model processed 5,034,263 input tokens and generated 618,284 output tokens, of which 518,080 were devoted to internal reasoning. The input tokens included the full-text PDFs of the citing and cited papers, which were provided to the model in each of the 90 stage-two runs. The total cost of all runs was approximately US $6.24, including about US $2.59 attributable to reasoning.

## S4 Motivation and mathematical specification of LPMs

To connect the qualitative patterns to experimental conditions, I estimated linear probability models (LPMs) for each of the 21 codes rather than logistic models. LPM coefficients remain on the probability scale and can be read directly as percentage point changes, which facilitates comparison across codes and conditions (Breen et al., 2018). While LPMs have familiar drawbacks, including heteroskedastic errors and fitted values that can fall outside [0,1], these are addressed here with robust standard errors and are secondary because the main quantities of interest are differences in outcome probabilities between experimental conditions.

Each model follows the balanced 2×3 design with six settings: two base prompts (*1-step* and *4-step*) crossed with three nudging conditions (*Toward, Away, No-nudge*). Models include the Base×Nudge interaction to allow nudge effects to differ across base prompts. I parameterize the model with *1-step*/*No-nudge* as the reference category, so the intercept equals the baseline probability of the code and all coefficients are interpretable as probability differences relative to that baseline.[4]

I summarize effects using average marginal effects (AMEs), which in an LPM are linear contrasts of coefficients and can be read directly as percentage point differences. I report three AMEs: AME(*4-step*), comparing *4-step* to *1-step* averaged over nudges, and AME(*Toward*) and AME(*Away*), comparing each nudge to *No-nudge* averaged over base prompts. In a balanced 2×3 design, these contrasts provide a compact summary aligned with the study's design-level questions while reducing the number of tests.

Formally, for each code $c$ and hypothesis $i$ the model is:

$$E[Y_{ci}] = \beta_0 + \beta_1 Base_i + \beta_2 Toward_i + \beta_3 Away_i + \beta_4 (Base_i \times Toward_i) + \beta_5 (Base_i \times Away_i) + \varepsilon_{ci}$$

where $Y_{ci}$ indicates whether a given code $c$ appears in hypothesis $i$ (1 = present, 0 = absent). $Base_i$ equals 1 for the *4-step* prompt and 0 for the *1-step* prompt. $Toward_i$ equals 1 for the *Toward* nudge and 0 otherwise, and $Away_i$ equals 1 for the *Away* nudge and 0 otherwise (so *No-nudge* is the omitted category). The interaction terms $(Base_i \times Toward_i)$ and $(Base_i \times Away_i)$ test whether the effects of nudging differ under the four-step prompt; their coefficients, $\beta_4$ and $\beta_5$ capture these differences. Finally, $\varepsilon_{ic}$ represents unexplained variation.

Standard errors are cluster-robust, clustered by run, because each run yields five hypotheses under the same prompt setting.

## S5 "Lexical echo": nudges measurably steer wording

To support my main analysis of nudging effects, I examined "lexical echo", that is, the model's reuse of vocabulary from the nudging paragraphs in the hypotheses. I lowercased the text, removed stopwords, and then constructed a set of 78 lexical items drawn from the two nudges. For some items I used stem forms such as "genealog" or "slippag". For others I kept the full word form where a stem would have made the feature too broad, for example by collapsing verbs and related grammatical forms into a single item when the noun use was analytically central, e.g. "reference" or "citation". For simplicity I refer to these lexical items as "stems" below. For each hypothesis–justification pair, I coded a binary indicator for the presence of each item and fitted the same 2×3 linear probability models as for the codes, with the same reference cell, interaction terms, clustering by run, and average marginal effects (AMEs). These models estimate changes in the model's probability of using nudge related vocabulary across the six experimental settings. Because testing many items raises the risk of false positives, I adjusted p-values for multiple comparisons using the Benjamini–Hochberg false discovery rate (FDR) procedure, which controls the expected share of false discoveries among rejected null hypotheses.

[4] Since each LPM uses a saturated 2×3 specification (an intercept plus five effects) that is algebraically equivalent to estimating a separate mean for each of the six settings, the fitted probabilities reproduce the observed cell averages. Formally, the OLS normal equations imply that residuals sum to zero within each cell, so the fitted cell mean equals the sample cell mean.

Two results stand out. First, the *Toward* nudge increased *Toward*-specific vocabulary in a focused way. On AME(*Toward*), seven *Toward* stems were higher under *Toward* than *No-nudge* after FDR correction: "enrol" rose by +18.0 pp, "genealog" by +13.3 pp, "slippag" by +12.0 pp, "refram" and "audience" by +9.3 pp each, and "rival" and "rewrit" increased by +5.3 pp and +4.7 pp. These terms align with the lineage, bridging, and shortcut motifs seeded in the *Toward* paragraph. By contrast, *Away*-specific stems did not survive FDR correction. Several showed small pre-FDR increases, for example "social" and "reader" (both +6.7 pp), and the *Toward* stem "audience" also rose under *Away* (+8.0), but none remained above the FDR-adjusted threshold relative to *No-nudge* after multiple-testing control.

Second, prompt structure moderated wording. On AME(*4-step*), the *4-step* prompt reduced reuse of broad citation terms that appear in both nudges. The probability of using "reference" was lower by about −35.6 pp, "citation" by −32.4 pp, and "source" by −8.9 pp relative to *1-step*. Several *Toward* stems also fell under *4-step*, including "down" (−6.2 pp), "downplay" (−4.9 pp), and "rival" (−3.6 pp). By contrast, "genealog" moved the other way under *4-step* (+3.6 pp) and is the only stem amplified by both base prompt structure and nudging. The stem "rival" shows the opposite cross-dimensional pattern, dropping under *4-step* despite rising under the *Toward* nudge.

These effects concern the model's probability of using nudge related vocabulary, not proof of direct copying. They show that examples can increase specificity yet also steer phrasing, and that scaffolded prompting can temper generic echo while leaving room for more analytic vocabulary. In this more fine grained sense, the lexical echo analysis complements the subtheme results and further addresses RQ3 by showing that prompt framing leaves measurable traces not only on which interpretations appear but also on the vocabulary through which they are articulated. Figure S5a plots all AMEs as dot-and-whisker estimates with 95% cluster-robust confidence intervals around a zero line, arranged in three panels for AME(*4-step*), AME(*Toward*), and AME(*Away*), with filled dots marking items with BH-FDR $q < 0.05$.

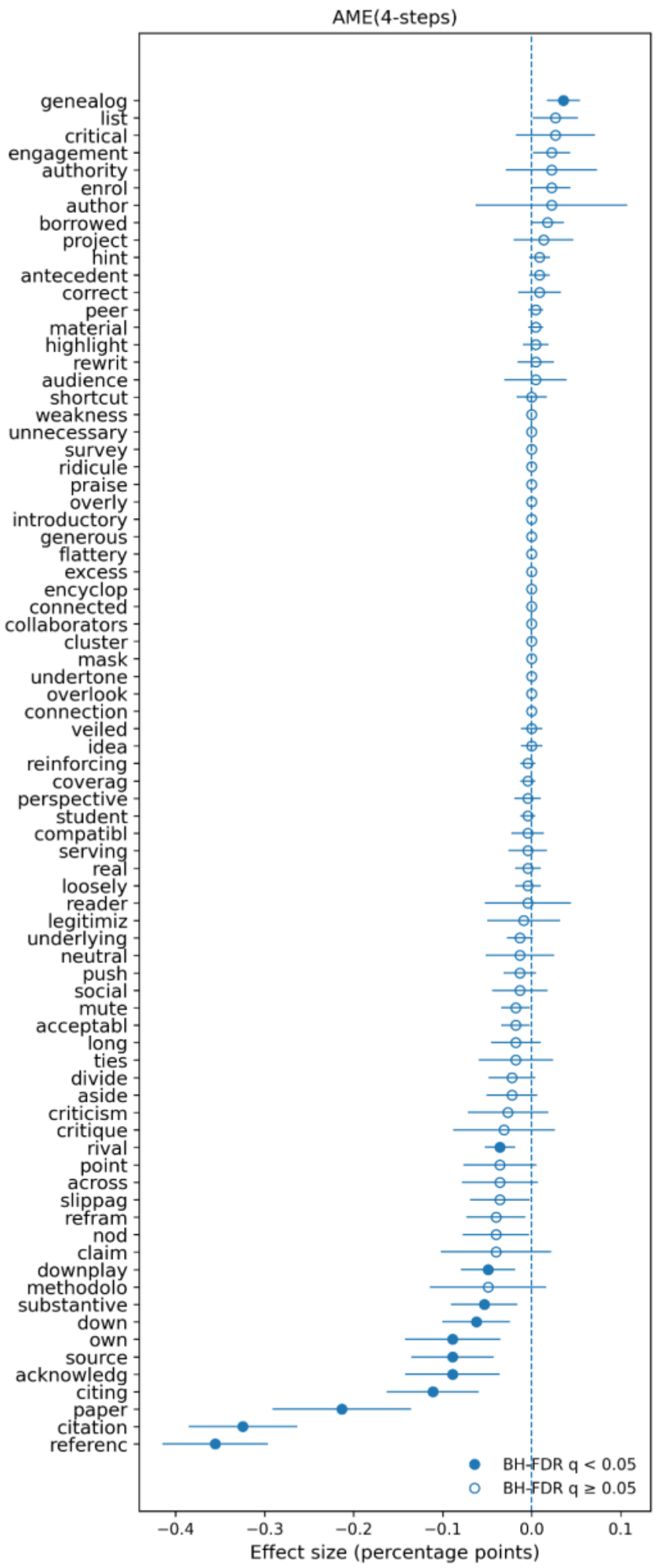

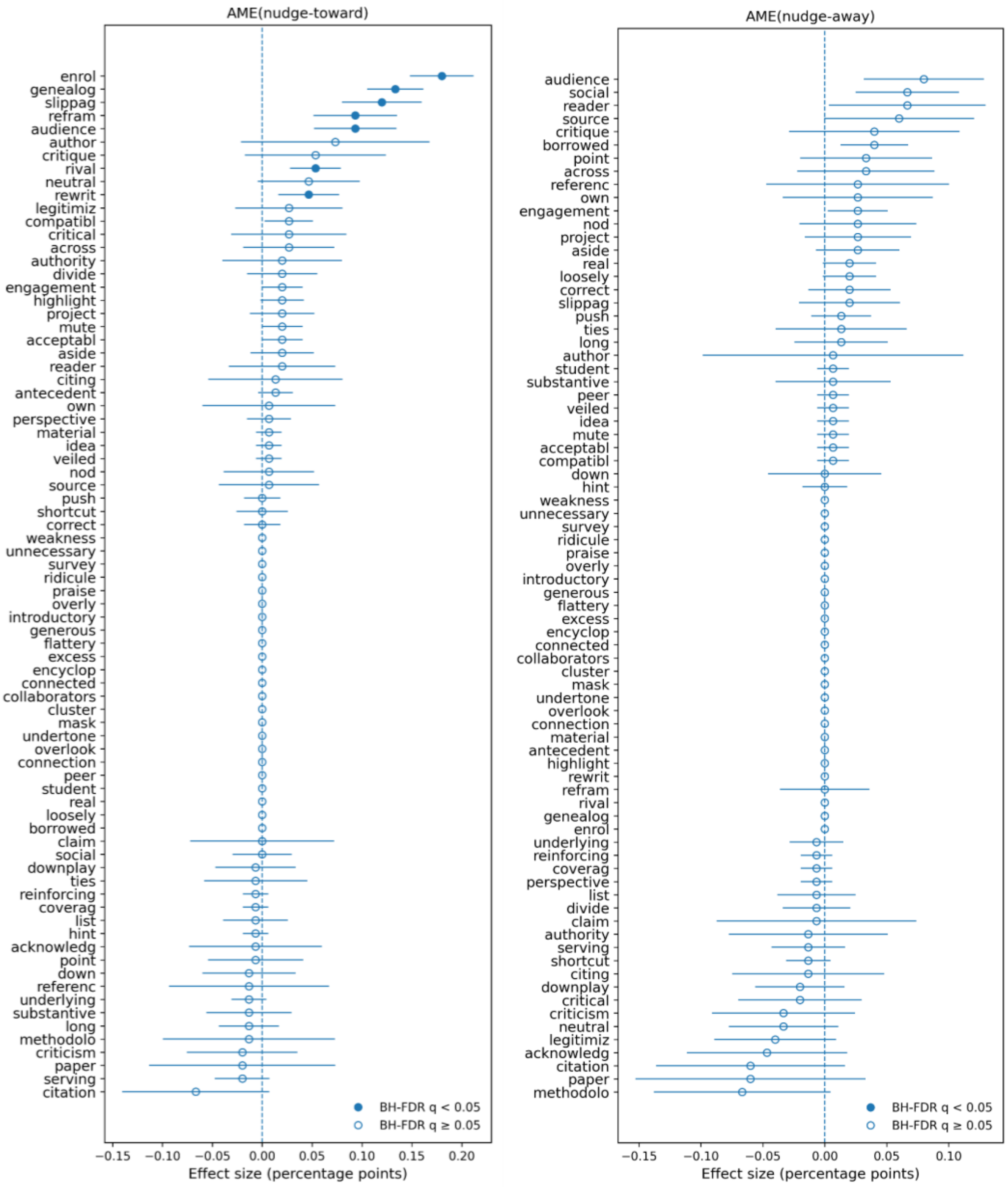

**Figure** S5a. Dot-and-whisker plots of average marginal effects (AMEs) for the 78 nudge-derived lexical items. Each subfigure shows one AME on the x-axis in percentage points, with one row per lexical item on the y-axis. Dots show AME point estimates and horizontal whiskers show unadjusted 95% cluster-robust confidence intervals, clustered by run. Filled dots indicate items with Benjamini–Hochberg FDR $q < 0.05$ within that subfigure, and open dots indicate $q \geq 0.05$. The vertical line at 0 indicates an effect size of 0. AME(*4-step*) compares *4-step* to *1-step*, averaged over the three nudges. AME(*Toward*) compares *Toward* to *No-nudge*, averaged over the two base prompts. AME(*Away*) compares *Away* to *No-nudge*, averaged over the two base prompts.

## Nudging markers

The following is a list of all nudging markers extracted using regular expressions from each of the examples provided in the two nudges

Examples and regular expressions from the *Toward* nudge:

**Example *Toward*-1**: “a neutral reference can mask a corrective undertone, hinting at overlooked connections or antecedents”

- [Nn]eutral
- [Rr]eferenc
- [Mm]ask
- [Cc]orrect
- [Uu]ndertone
- [Hh]int
- [Oo]verlook
- [Cc]onnection
- [Aa]ntecedent

**Example *Toward*-2**: “By highlighting some works and downplaying others, a paper can rewrite an idea’s genealogy”

- [Hh]ighlighting
- [Ww]ork
- [Dd]ownplay
- [Dd]own
- [Pp]aper
- [Rr]ewrit
- [Ii]dea
- [Gg]enealog

**Example *Toward*-3**: “A nod to a critical source may mute that critique, acknowledging it only to push it aside”

- [Nn]od
- [Cc]ritical
- [Ss]ource
- [Mm]ute
- [Cc]ritique
- [Aa]cknowledg
- [Pp]ush
- [Aa]side

**Example *Toward*-4**: “Citations can also legitimize shortcuts, suggesting that methodological slippage is acceptable because others have done it”

- [Cc]itation
- [Ll]egitimiz
- [Ss]hortcut

- [Mm]ethodolo
- [Ss]lippag
- [Aa]cceptabl

**Example *Toward*-5**: “by citing across divides, authors may enroll rival audiences while reframing their perspectives as compatible with their own”

- [Cc]iting
- [Aa]cross
- [Dd]ivide
- [Aa]uthor # NOTE: matches r[Aa]uthority too
- [Ee]nrol
- [Rr]ival
- [Aa]udience
- [Rr]efram
- [Pp]erspective
- [Cc]ompatibl
- [Oo]wn

Examples and regular expressions from the *Away* nudge

**Example *Away*-1**: “overly generous praise can mask underlying criticism, serving as a veiled form of ridicule, where excess flattery highlights the weakness of the work”

- [Oo]verly
- [Gg]enerous
- [Pp]raise
- [Mm]ask
- [Uu]nderlying
- [Cc]riticism
- [Ss]erving
- [Vv]eiled
- [Rr]idicule
- [Ee]xcess
- [Ff]lattery
- [Hh]ighlight
- [Ww]eakness
- [Ww]ork

**Example *Away*-2**: “An otherwise unnecessary reference may act as a nod to collaborators, reinforcing social ties rather than substantive claims”

- [Uu]nnecessary
- [Rr]eferenc
- [Nn]od
- [Cc]ollaborators
- [Rr]einforcing
- [Ss]ocial

- [Tt]ies
- [Ss]ocial [Tt]ies
- [Ss]ubstantive
- [Cc]laim

**Example *Away*-3**: “A cluster of citations can lend borrowed authority even when the sources are only loosely connected”

- [Cc]luster
- [Cc]itation
- [Bb]orrowed
- [Aa]uthority
- [Ss]ource
- [Ll]oosely
- [Cc]onnected

**Example *Away*-4**: “Long reference lists may project encyclopedic coverage without real engagement”

- [Ll]ong
- [Rr]eferenc
- [Ll]ist
- [Rr]eference [Ll]ist
- [Pp]roject
- [Ee]ncyclop
- [Cc]overag
- [Rr]eal
- [Ee]ngagement

**Example *Away*-5**: “And some references work less for peers than for students, pointing readers toward introductory or survey material”

- [Rr]eferenc
- [Pp]eer
- [Ss]tudent
- [Pp]oint
- [Rr]eader
- [Ii]ntroductory
- [Ss]urvey
- [Mm]aterial

## S6 Analysis of the “reiter” cue

The overall tendency to associate “reiterated” with muted lineage and positioning readings also appears when we move from the *4-step* prompts to the full set of runs and look at co-occurrence patterns for the stem of the “reiter” cue. Up to this point I have drawn only on the *4-step* settings, where expectation checks, cue analyses, and extended context analyses provide an explicit intermediate trace. The *1-step* prompts do not have such intermediary steps, so for those runs the final hypotheses are the only available material. If we count how often each of the 21 codes co-occurs with the stem “reiter” in a hypothesis, similar patterns appear under both base prompts. In *4-step* runs, codes whose occurrences most often include “reiter” are *MuteGW*, with 94% of its occurrences co-occurring with “reiter”, *Canon*, 80%, *MuteP*, 67%, *UseGW*, 65%, *PrioP*, 63%, and *SideP*, 59%. In *1-step* runs, the same cluster appears, with *MuteGW*, 95%, *UseGW*, 80%, *PrioP*, 77.8%, *Canon*, 71%, and *SideP*, 67%, more likely to occur in hypotheses that mention “reiter” than in those that do not. A similar picture emerges across the three nudging settings. Under *Toward*, *Away*, and *No-nudge*, the same core cluster of codes is again most likely to co-occur with “reiter”, namely *MuteGW*, *Canon*, *UseGW*, *PrioP*, and *SideP*, with *MuteP* joining the cluster under the Away nudge and only small differences in their relative frequencies.

An additional LPM with a binary indicator for the presence of “reiter” as the outcome shows that prompt settings do influence how often the model uses this cue. In the reference setting, the probability that a hypothesis contains “reiter” is about one quarter, and it ranges from 21% to 35% across the six cells, with the highest value under *4-step*/*Toward* and the lowest under *1-step*/*Away*. The average marginal effects for base prompt and nudging are modest, ranging from −2 pp for AME(*Away*) to +5 pp for AME(*Toward*), and their 95% confidence intervals include 0. By contrast, two cell-level contrasts involving *4-step*/*Toward* have 95% confidence intervals that exclude 0 (*4-step*/*Toward* vs *4-step*/*No-nudge*, +11 pp, and *4-step*/*Toward* vs *1-step*/*Away*, +13 pp). Overall, the LPM points to a small shift toward explicit “reiter” use in the scaffolded *Toward* setting, while the basic co-occurrence cluster around that cue remains broadly similar across prompt structures and nudges.

## S7 Should G&W have credited Price for the ref–cit distinction?

The following passage reproduces a conversation with ChatGPT-5 in “extended thinking” mode about whether Gilbert and Woolgar (1974) should have cited Price (1970) when introducing the ref–cit distinction. It follows on from an earlier, unquoted exchange in which I used the same interface to revise specific passages of the then current draft of this paper. ChatGPT’s answers here largely confirmed my own judgement after reading both papers, but I include it because of the way it marshals textual evidence and articulates competing credit intuitions. The model independently identified the close similarity between Price’s and G&W’s formulations of the distinction, Price’s explicit presentation of it as his own terminological proposal, and the fact that G&W were clearly familiar with Price’s paper. Although the main experiments in this article use GPT-5 via the API, this short browser based exchange, together with the earlier drafting conversation, points in the same direction and illustrates how an LLM can support interpretative CCA by assembling cues across texts and staging alternative readings without claiming to resolve authorial intent.

**ME:** Ok, something else now. Nowhere in the paper I have so far established the plausibility of Gilbert's interpretation that C&M meant footnote 6 as an admonishment about credit. The question is: How would we judge the situation if we read both Price and G&W. Do you think, this is a fair question to pose and answer somewhere in the paper?

**ChatGPT:** Yes, I think that is a very fair question to pose, and it fits nicely with what the paper is already doing.
Right now you treat Gilbert's admonishment reading mainly as a historical and methodological pivot. You show that GPT-5 picks up the same cues but channels them into other readings. What you have not yet done is step back and say, from the position of a well informed human reader who has both Price and G&W in front of them, how strong is Gilbert's own interpretation as one possible reading among others.
That kind of move would help you in at least three ways:

1. It clarifies what "success" on the Gilbert test would even mean
   At the moment RQ4 is model focused. You ask how close GPT-5 comes to Gilbert's reading, but you do not really tell the reader whether Gilbert's reading is an obvious, middling, or fairly speculative interpretation once you actually look at the two cited texts.
   A short human-facing check like "if we read Price and G&W ourselves, how natural does the admonishment reading feel relative to other options" gives the reader a sense of what bar the model is being compared against.
2. It reinforces your "multiple plausible readings" stance, not intention hunting
   The danger would be if this slipped into "deciding what C&M really meant." You can avoid that by framing it explicitly as a thought experiment about plausibility for a well informed reader, not as a verdict on author intention.
   For example, you could say something like:
   – Given Price's and G&W's own handling of the ref / cite distinction and the absence or presence of cross citation, a reader who knows both texts can see why Gilbert's admonishment reading is available.
   – At the same time, a neutral lineage reading remains viable, because the wording and tone of footnote 6 do not force the reader into an accusatory interpretation.
   That lets you stay aligned with the rest of the paper, where the emphasis is on interpretive underdetermination and the production of plausible alternatives, rather than on recovering a single correct label.
3. It gives you a human baseline to set next to the model's repertoire
   You already show that GPT-5 gets very close in terms of ingredients: it notices that G&W do not credit Price for the distinction, it latches onto "introduced… then reiterated," it sees how Price recurs and G&W fade.
   If you add a brief "from the standpoint of my own rereading of Price and G&W, here is where I would place Gilbert's interpretation in the space of plausible readings," then the comparison becomes richer:
   – The human scholar's judgement about plausibility
   – The model's distribution of hypotheses
   – Gilbert's original 1977 reading
   That triangulation really suits a paper about LLMs as co analysts rather than as oracles.

In terms of where to put this, two natural spots:

- Near the end of “Footnote 6 as a hard case,” as a short paragraph that says: if we actually read Price and G&W, how does footnote 6 look. That would anchor the case historically and textually before you turn to models.
- Or inside the RQ4 “Gilbert test” subsection, perhaps just after you note that none of the 450 hypotheses reproduces the admonishment reading in full. There you could say “to see what is at stake, it is helpful to ask how a human reader might judge footnote 6 after reading Price and G&W side by side.”

The one thing I would avoid is overclaiming. I would not say “a fair reading shows that Gilbert was right” or “Gilbert overread the case.” Instead I would situate your judgement with hedges that match your epistemic stance elsewhere. For instance:

- “To me, having reread both texts, Gilbert’s admonishment reading is plausible but not compelled.”
- Or, “The reading becomes more salient once one knows that G&W do not credit Price, yet a neutral genealogy remains available even to a well informed reader.”

That keeps the question fair, acknowledges the work of actually reading Price and G&W, and folds the answer back into your broader claim that interpretative CCA is about mapping and comparing readings, not resolving them once and for all.

**ME:** Ok, here are the two papers in question.[5] Tell me what you think? Should G&W have cited Price where they talk about the ref-cit distinction? IF yes, why? If no, why not?

**ChatGPT:** Short answer: on balance, yes, I think they should have cited Price at the point where they introduce the ref–cit distinction, and I can also see why a contemporary reader might not have experienced the omission as egregious.

First, the textual evidence.

In the 1970 chapter, Price explicitly frames the distinction as something he is coining:

“It seems to me a great pity to waste a good technical term by using the words citation and reference interchangeably. I, therefore, propose and adopt the convention that if Paper R contains a bibliographic footnote using and describing Paper C, then R contains a reference to C, and C has a citation from R […]” (Price, 1970)

He marks this as a proposal, uses R and C, and ties the distinction directly to his larger argument about using citation counts as unobtrusive measures of cumulation and “scholarliness.” The definition is not just housekeeping, it underpins the whole piece.

Gilbert and Woolgar, four years later, write:

“In order to avoid confusion in describing this literature, we must distinguish carefully between a reference and a citation. A paper (R) may give a reference to an earlier paper (C). In so doing, a paper (C) is given a citation by the later paper (R).” (Gilbert and Woolgar, 1974)

The structure and even the R/C notation closely mirror Price. They then criticise other authors for not keeping the terms apart, and proceed to build a chunk of their review on that distinction.

Crucially, it is not that they were unaware of the 1970 piece. In the very same section they later cite “Citation measures compared with hard science, soft science, technology and nonscience” for the immediacy effect and Price’s Index, and they discuss that index at some length. So they clearly had the text in front of them and treated other ingredients in it as attributable contributions.

---

[5] I uploaded the full-text PDFs of both Price (1970) and Gilbert and Woolgar (1974) together with this prompt.

Given that combination – Price explicitly "proposing" the distinction, G&W reproducing both the content and notation, and their own paper positioning itself as a critical review of the quantitative literature – I think the fair-credit move would have been a small clause such as "following Price (1970), we distinguish between…" at the point where they define the terms. That would have:

- Acknowledged that the specific way of carving up "reference" and "citation" is not theirs.
- Linked their criticism of others' terminology to an existing attempt to stabilise it, instead of implicitly presenting themselves as the first to insist on the distinction.
- Made it easier for later readers to trace the genealogy of a convention that did, in fact, become standard in bibliometrics.

That is the "yes, they should have" side.

There is also a "why someone in 1974 might not have felt compelled to" story:

- The distinction is quite intuitive once you are working with citation indexes, and there is a sense in which it can feel like a piece of tidying-up prose rather than a substantive claim.
- Terminological conventions of this kind often circulate informally. By the time G&W wrote, the ref–cit distinction may already have been in their mental toolbox as "how people talk in this area" rather than "Price's invention," even though it does trace back to his footnote.
- They do credit Price heavily elsewhere in the paper for more central points: the exponential growth of science, the logistic-growth claim, and especially the immediacy effect and Price's Index. From inside that texture of repeated Price citations, the missing reference at the definitional sentence can look like a minor oversight rather than a deliberate erasure.

So if you ask "is it understandable that they did not cite him there?", I would say yes. If you ask "given our usual norms about attributing technical distinctions, especially when you reuse both the content and the notation and you have the source on your desk, would a citation have been appropriate?", I would also say yes.

That is why Gilbert's later suggestion that Price "introduced" and G&W "reiterated" the distinction is not crazy as a matter of credit. Your own question then becomes whether C&M's wording carries a hint of "and they really should have said so at the time." The textual set-up certainly makes that reading available, even if it does not force it.